\documentclass{article}
\usepackage{ijcai25}
\usepackage{times}
\usepackage{soul}
\usepackage{url}
\usepackage[hidelinks]{hyperref}
\usepackage[utf8]{inputenc}
\usepackage[small]{caption}
\usepackage{graphicx}
\usepackage{amsmath}
\usepackage{amsthm}
\usepackage{booktabs}
\usepackage{amsmath,amsfonts,bm}
\usepackage{subfigure}
\usepackage[table]{xcolor}
\definecolor{lightgray}{gray}{0.95}
\usepackage{multirow} 
\usepackage{algorithm}
\usepackage{algorithmic}
\usepackage[switch]{lineno}
\newcommand{\ie}{\textit{i.e.}}
\newcommand{\eg}{\textit{e.g.}}
\title{Zero-shot Generalist Graph Anomaly Detection with Unified Neighborhood Prompts}

\author{
Chaoxi Niu$^{1 \dagger}$
\and
Hezhe Qiao$^{2 \dagger} $\and
Changlu Chen$^{3}$\and
Ling Chen$^{1}$\And
Guansong Pang$^{2}$\thanks{Corresponding author: Guansong Pang (gspang@smu.edu.sg), $^\dagger$ Co-first author.}\\
\affiliations
$^{1}$ AAII, University of Technology Sydney, Sydney, Australia\\
$^{2}$ School of Computing and Information Systems, Singapore Management University, Singapore\\
$^{3}$ Faculty of Data Science, City University of Macau, Macau, China\\
\emails
{chaoxi.niu@student.uts.edu.au,
hezheqiao.2022@phdcs.smu.edu.sg,
clchen@cityu.edu.mo,
ling.chen@uts.edu.au,
gspang@smu.edu.sg}
}

\begin{document}

\maketitle

\begin{abstract}
Graph anomaly detection (GAD), which aims to identify nodes in a graph that significantly deviate from normal patterns, plays a crucial role in broad application domains. However, existing GAD methods are one-model-for-one-dataset approaches, \ie, training a separate model for each graph dataset. This largely limits their applicability in real-world scenarios. To overcome this limitation, we propose a novel zero-shot generalist GAD approach \textbf{UNPrompt} that trains a one-for-all detection model, requiring the training of one GAD model on a single graph dataset and then effectively generalizing to detect anomalies in other graph datasets without any retraining or fine-tuning. The key insight in UNPrompt is that i) the predictability of latent node attributes can serve as a generalized anomaly measure and ii) generalized normal and abnormal graph patterns can be learned via latent node attribute prediction in a properly normalized node attribute space. UNPrompt achieves a generalist mode for GAD through two main modules: one module aligns the dimensionality and semantics of node attributes across different graphs via coordinate-wise normalization, while another module learns generalized neighborhood prompts that support the use of latent node attribute predictability as an anomaly score across different datasets. Extensive experiments on real-world GAD datasets show that UNPrompt significantly outperforms diverse competing methods under the generalist GAD setting, and it also has strong superiority under the one-model-for-one-dataset setting. Code is available at \url{https://github.com/mala-lab/UNPrompt}.
\end{abstract}

\section{Introduction}
Graph anomaly detection (GAD) has attracted extensive research attention in recent years \cite{pang2021deep,qiao2024deep} due to the broad applications in various domains such as spam review detection in online shopping networks \cite{mcauley2013amateurs,rayana2015collective} and malicious user detection in social networks \cite{yang2019mining}. To handle high-dimensional node attributes and complex structural relations between nodes, graph neural networks (GNNs) \cite{kipf2016semi,wu2020comprehensive} have been widely exploited for GAD due to their strong ability to integrate the node attributes and graph structures. These methods can be roughly divided into two categories, \ie, supervised and unsupervised methods. The former formulates GAD as a binary classification problem and aims to capture anomaly patterns under the guidance of labels \cite{tang2022rethinking,peng2018anomalous,gao2023addressing}. By contrast, due to the difficulty of obtaining these class labels, the latter category takes the unsupervised approach that aims to learn normal graph patterns, \eg, via data reconstruction or other proxy learning tasks that are related to GAD \cite{qiao2024truncated,liu2021anomaly,ding2019deep,huang2022hop}. 
\begin{figure*}
\centering
\includegraphics[width=0.748\linewidth]{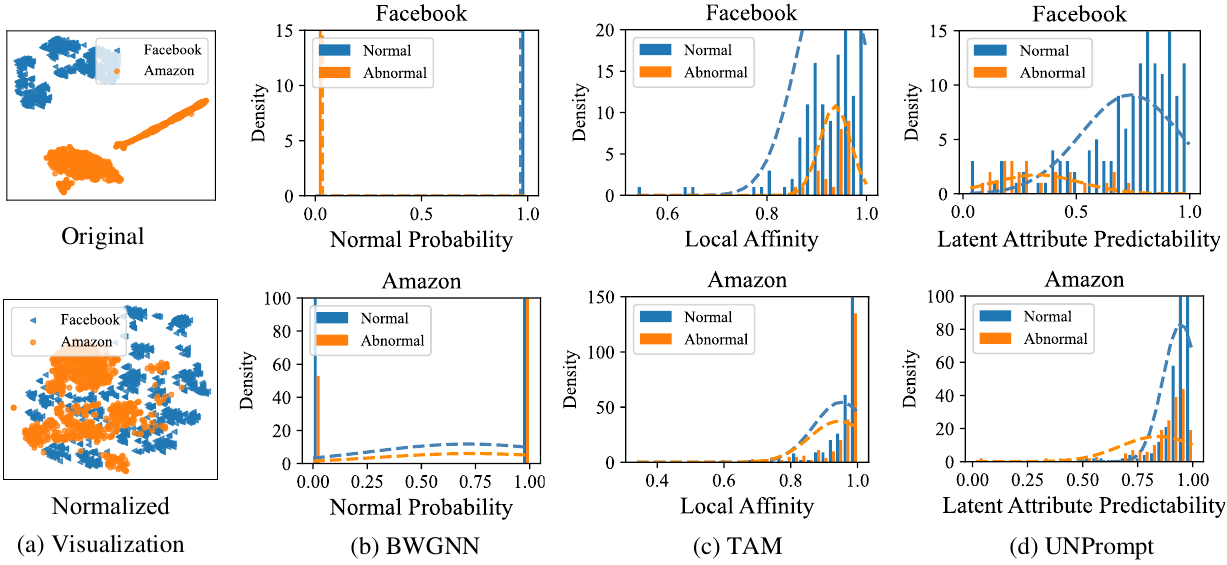}
\caption{\textbf{(a)} Visualization of two popular GAD datasets: Facebook and Amazon, where the node attributes are unified into a common semantic space via our proposed normalization compared to the original heterogeneous raw attributes. \textbf{(b)-(d)} The anomaly scores of BWGNN (normal probability), TAM (local affinity), and UNPrompt (latent attribute predictability) on the two datasets, where the methods are all trained on Facebook and tested on Amazon under the zero-shot setting. It is clear that BWGNN and TAM struggle to generalize from Facebook to Amazon, while UNPrompt can learn well to generalize across the datasets.}
\label{motivation}
\end{figure*}

Despite their remarkable detection performance, these methods need to train a dataset-specific GAD model for each graph dataset. This one-model-for-one-dataset paradigm limits their applicability in real-world scenarios since training a model from scratch can incur extensive computation costs and require a large amount of labeled data for supervised GAD methods on each dataset \cite{liu2024arc,qiao2024deep}. Training on a target graph may not even be possible due to data privacy protection and regulation. To address this limitation, a new one-for-all anomaly detection (AD) paradigm, called generalist anomaly detection \cite{zhu2024toward,zhou2024anomalyclip}, has been proposed for image AD with the emergence of foundation models such as CLIP \cite{radford2021learning}. This new direction aims to learn a generalist detection model on auxiliary datasets so that it can generalize to detect anomalies effectively in diverse target datasets without any re-training or fine-tuning. This paper explores this direction in the area of GAD.

Compared to image AD, there are some unique challenges for learning generalist models for GAD. First, unlike image data where raw features are in the same RGB space, the node attributes of graphs from different applications and domains can differ significantly in node attribute dimensionality and semantics. For example, as a shopping network dataset, Amazon contains the relationships between users and reviews, and the node attribute dimensionality is 25. Differently, Facebook, a social network dataset, describes relationships between users with 576-dimensional attributes. Second, generalist AD models on image data rely on the superior generalizability learned in large visual-language models (VLMs) through pre-training on web-scale image-text-aligned data \cite{zhu2024toward,zhou2024anomalyclip}, whereas there are no such foundation models for graphs \cite{liu2023towards}. Therefore, the key question here is: \textit{can we learn generalist models for GAD on graph data with heterogeneous node attributes and structure without the support of foundation models?}

To address these challenges, we propose \textbf{UNPrompt}, a novel generalist GAD approach that learns \textit{Unified Neighborhood Prompts} on a single auxiliary graph dataset and then effectively generalizes to directly detect anomalies in other graph datasets under a \textbf{zero-shot} setting. The key insight in UNPrompt is that i) the predictability of latent node attributes can serve as a generalized anomaly measure and ii) graph-agnostic normal and abnormal patterns can be learned via latent node attribute prediction in a properly normalized node attribute space. UNPrompt achieves this through two main modules, including \textit{coordinate-wise normalization-based node attribute unification} and \textit{neighborhood prompt learning}. The former module aligns the dimensionality of node attributes across graphs and transforms the semantics into a common space via coordinate-wise normalization, as shown in Figure~\ref{motivation}(a). In this way, the diverse distributions of node attributes are calibrated into the same semantic space. On the other hand, the latter module focuses on modeling graph-agnostic normal and abnormal patterns across different graph datasets. This is achieved in UNPrompt by learning generalized graph prompts in the normalized attributes of the neighbors of a target node via a latent node attribute prediction task. In doing so, UNPrompt, being trained on a single graph with a small GNN, can yield effective anomaly scores for detecting anomalous nodes in diverse unseen graphs without any re-training, as shown in Figure~\ref{motivation}(b)-(d).

Overall, the main contributions of this paper are summarized as follows. \textbf{(1)} We propose a novel zero-shot generalist GAD approach, UNPrompt. To the best of our knowledge, this is the first method that exhibits effective zero-shot GAD performance across various graph datasets. There is concurrent work on generalist GAD \cite{liu2024arc}, but it can only work under a few-shot setting. \textbf{(2)} By unifying the heterogeneous distributions in the node attributes across different graphs, we further introduce a novel neighborhood prompt learning module that utilizes a neighborhood-based latent node attribute prediction task to learn generalized prompts. This enables the zero-shot GAD of UNPrompt across different graphs. \textbf{(3)} Extensive experiments on real-world GAD datasets show that UNPrompt significantly outperforms state-of-the-art competing methods under the zero-shot generalist GAD. \textbf{(4)} We show that UNPrompt can also work in the conventional one-model-for-one-dataset setting, outperforming state-of-the-art models in this popular GAD setting.

\section{Related Work}
\noindent \textbf{Graph Anomaly Detection.}
Existing GAD methods can be roughly categorized into unsupervised and supervised approaches \cite{qiao2024deep}. The unsupervised methods are typically built using data reconstruction, self-supervised learning, and learnable graph anomaly measures \cite{qiao2024deep,liu2022bond}. The reconstruction-based approaches like DOMINANT \cite{ding2019deep} and AnomalyDAE \cite{fan2020anomalydae} aim to capture the normal patterns in the graph, where the reconstruction error in both graph structure and attributes is utilized as the anomaly score. CoLA \cite{liu2021anomaly} and SL-GAD \cite{zheng2021generative} are representative self-supervised learning methods assuming that normality is reflected in the relationship between the target node and its contextual nodes. The graph anomaly measure methods typically leverage the graph structure-aware anomaly measures to learn intrinsic normal patterns for GAD, such as node affinity in TAM \cite{qiao2024truncated}.
In contrast to the unsupervised approaches, the supervised approaches have shown substantially better detection performance in recent years due to the incorporation of labeled anomaly data \cite{liu2021pick,chai2022can}. Most supervised methods concentrate on the design of propagation mechanisms and spectral feature transformations to address the notorious over-smoothing issues \cite{tang2022rethinking,gao2023addressing,chai2022can}. Although both approaches can be adapted for zero-shot GAD by directly applying the trained GAD models to the target datasets, they struggle to capture generalized normal and abnormal patterns for GAD across different graph datasets. There are some works on cross-domain GAD \cite{ding2021cross,wang2023cross} that aim to transfer knowledge from a labeled graph dataset to a target dataset, but it is a fundamentally different problem from generalist GAD since cross-domain GAD requires training on both source and target graph datasets.

\noindent \textbf{Graph Prompt Learning.} 
Prompt learning, initially developed in natural language processing, seeks to adapt large-scale pre-trained models to different downstream tasks by incorporating learnable prompts while keeping the pre-trained models frozen \cite{liu2023pre}. Specifically, it designs task-specific prompts capturing the knowledge of the corresponding tasks and enhances the compatibility between inputs and pre-trained models to improve the pre-trained models in downstream tasks. Recently, prompt learning has been explored in graphs to unify multiple graph tasks \cite{sun2023all,liu2023graphprompt} or improve the transferability of graph models on the datasets across the different domains \cite{li2024zerog,zhao2024all} which optimize the prompts with labeled data of various downstream tasks \cite{fang2024universal,liu2023graphprompt}. Although being effective in popular graph learning tasks like node classification and link prediction, they are inapplicable to generalist GAD due to the unsupervised nature and/or irregular distributions of anomalies.

\noindent\textbf{Generalist Anomaly Detection.} 
Generalist AD has been very recently emerging as a promising solution to tackle sample efficiency and model generalization problems in AD. There have been a few studies on non-graph data that have large pre-trained models to support the generalized pattern learning, such as image generalist AD \cite{zhou2024anomalyclip,zhu2024toward}. However, it is a very challenging task for graph data due to the lack of such pre-trained models. Recently, a concurrent approach, ARC \cite{liu2024arc}, introduces an effective framework that leverages in-context learning to achieve generalist GAD without relying on large pre-trained GNNs. Unlike ARC which focuses on a few-shot GAD setting, \ie, requiring the availability of some labeled nodes in the target testing graph dataset, we tackle a zero-shot GAD setting, assuming no access to any labeled data during inference stages.

\section{Methodology}

\subsection{Preliminaries}
\paragraph{Notations.}
Let $\mathcal{G} = (\mathcal{V}, \mathcal{E})$ be an attributed graph with $N$ nodes, where $\mathcal{V} = \{v_1, v_2, \ldots, v_N\}$ represents the node set and $\mathcal{E}$ is the edge set. The attributes of nodes can be denoted as $X = [\mathbf{x}_1, \mathbf{x}_2, \ldots, \mathbf{x}_N]^T\in\mathbb{R}^{N\times d}$ and the edges between nodes can be presented by an adjacency matrix $A\in\{0,1\}^{N\times N}$ with $A_{ij} = 1$ if there is an edge between $v_i$ and $v_j$ and $A_{ij} = 0$ otherwise. For simplicity, the graph can be represented as $\mathcal{G} = (A, X)$. In GAD, the node set can be divided into a set of normal nodes $\mathcal{V}_{n}$ and a set of anomalous nodes $\mathcal{V}_{a}$. Typically, the number of normal nodes is significantly larger than the anomalous nodes, \ie, $|\mathcal{V}_n| \gg |\mathcal{V}_a|$. Moreover, the anomaly labels can be denoted as $\mathbf{y}\in\{0,1\}^N$ with $\mathbf{y}_i = 1$ if $v_i\in\mathcal{V}_a$ and $\mathbf{y}_i = 0$ otherwise.

\paragraph{Conventional GAD.} Conventional GAD typically focuses on model training and anomaly detection on the same graph. Specifically, given a graph $\mathcal{G}$, an anomaly scoring model $f:\mathcal{G}\rightarrow \mathbb{R}$ is optimized on $\mathcal{G}$ in a supervised or unsupervised manner. Then, the model is used to detect anomalies within the same graph. The model is expected to generate higher anomaly scores for abnormal nodes than normal nodes, \ie, $f(v_i) < f(v_j)$ if $v_i \in \mathcal{V}_n$ and $v_j\in \mathcal{V}_a$.

\paragraph{Generalist GAD.}
Generalist GAD aims to learn a generalist model $f$ on a single training graph so that $f$ can be directly applied to different target graphs across diverse domains without any fine-tuning or re-training. More specifically, the model is optimized on $\mathcal{G}_{\text{train}}$ with the corresponding anomaly labels $\mathbf{y}_{\text{train}}$. After model optimization, the learned $f$ is utilized to detect anomalies within different unseen target graphs $\mathcal{T}_{\text{test}} = \{\mathcal{G}^{(1)}_{\text{test}}, \ldots, \mathcal{G}^{(n)}_{\text{test}}\}$ which has heterogeneous attributes and/or structures to $\mathcal{G}_{\text{train}}$, \ie, $\mathcal{G}_{\text{train}} \cap \mathcal{T}_{\text{test}} = \emptyset$. Depending on whether labeled nodes of the target graph are provided during inference, the generalist GAD problem can be further divided into two categories, \ie, \textbf{few-shot} and \textbf{zero-shot} settings. We focus on the zero-shot setting where the generalist models cannot get access to any labeled data of the testing graphs during both training and inference.

\subsection{Overview of UNPrompt}
\begin{figure*}
\centering
\includegraphics[width=0.9\textwidth]{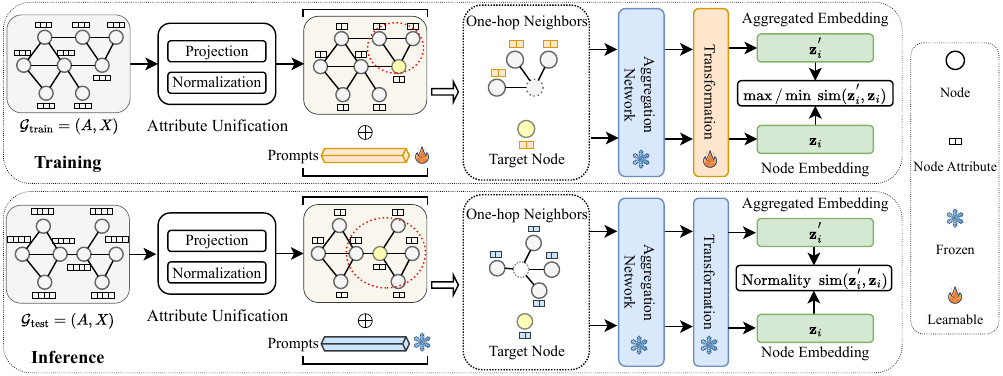}
\caption{Overview of UNPrompt. During training, the neighborhood prompts are optimized to capture generalized patterns by maximizing the predictability of the latent attributes of normal nodes while minimizing that of abnormal nodes. At the inference stage, the learned prompts are directly applied to the testing nodes, and the latent attribute predictability is used for GAD.}
\label{framework}
\end{figure*}

The framework of UNPrompt is illustrated in Figure~\ref{framework}, which consists of two main modules, coordinate-wise normalization-based node attribute unification and neighborhood prompt learning. For all graphs, the attribute unification aligns the dimensionality of node attributes and transforms the semantics into a common space. Then, in the normalized space, the generalized latent attribute prediction task is performed with the neighborhood prompts to learn generalized GAD patterns. Specifically, UNPrompt aims to maximize the predictability of the latent attributes of normal nodes while minimizing those of abnormal nodes. In this paper, we evaluate the predictability via the similarity. In doing so, the graph-agnostic normal and abnormal patterns are incorporated into the prompts. During inference, the target graph is directly fed into the learned models after node attribute unification without any re-training or labeled nodes of the graph. For each node, the predictability of latent node attributes is directly used as the normal score for final anomaly detection. 

\subsection{Node Attribute Unification}\label{unification}
Graphs from different distributions and domains significantly differ in the dimensionality and semantics of node attributes. The premise of developing a generalist GAD model is to unify the dimensionality and semantics of node attributes into the same space. To address this issue, we propose a simple yet effective node attribute unification module that consists of feature projection and coordinate-wise normalization.

\paragraph{Feature Projection.} 
To address the inconsistent attribute dimensions across graphs, various feature projection methods can be utilized, such as singular value decomposition \cite{stewart1993early} (SVD) and principal component analysis \cite{abdi2010principal} (PCA). Formally, given the attribute matrix $X^{(i)}\in\mathbb{R}^{N^{(i)}\times d^{(i)}}$ of any graph $\mathcal{G}^{(i)}$ from $\mathcal{G}_{\text{train}}\cup \mathcal{T}_{\text{test}}$, we transform it into $\tilde{X}^{(i)}\in \mathbb{R}^{N^{(i)}\times d^{'}}$ with the common dimensionality of $d^{'}$,
\begin{equation}
X^{(i)}\in\mathbb{R}^{N^{(i)}\times d^{(i)}} \xrightarrow[\text{Projection}]{\text{Feature}} \tilde{X}^{(i)}\in \mathbb{R}^{N^{(i)}\times d^{'}}\, .
\end{equation}

\paragraph{Coordinate-wise Normalization.}
Besides the inconsistent attribute dimensionality, the divergent node attribute semantics and distributions of different graphs pose significant challenges for generalist GAD. A recent study \cite{li2024zero} has demonstrated that semantic differences across datasets are mainly reflected in the distribution shifts and calibrating the distributions into a common frame helps learn more generalized AD models. Inspired by this, we propose to use coordinate-wise normalization to align the semantics and unify the distributions across graphs. Specifically, the transformed attribute matrix $\tilde{X}^{(i)}$ is shifted and re-scaled to have mean zeros and variance ones via the following equation:
\begin{equation}
\bar{X}^{(i)} = \frac{\tilde{X}^{(i)} - \boldsymbol{\mu}^{(i)}}{\boldsymbol{\sigma}^{(i)}}\, ,
\end{equation}
where $\boldsymbol{\mu}^{(i)} = [{\mu}^{(i)}_1, \ldots, {\mu}^{(i)}_{d^{'}}]$ and $\boldsymbol{\sigma}^{(i)} = [{\sigma}^{(i)}_1, \ldots, {\sigma}^{(i)}_{d^{'}}]$ are the coordinate-wise mean and variance of $\tilde{X}^{(i)}$ of the graph $\mathcal{G}^{(i)}$. In this way, the distributions of normalized attributes along each dimension are the same within and across graphs, as shown in Figure \ref{motivation}(a). 
This helps to capture the generalized normal and abnormal patterns for generalist GAD. 

\subsection{Neighborhood Prompt Learning}

\paragraph{Latent Node Attribute Predictability as Anomaly Score.}
In this paper, we reveal that the predictability of latent node attributes can serve as a generalized anomaly measure, and highly generalized normal and abnormal graph patterns can be learned via latent node attribute prediction in the normalized attribute space with neighborhood prompts. The key intuition of this anomaly measure is that normal nodes tend to have more connections with normal nodes of similar attributes due to prevalent graph homophily relations, resulting in a more homogeneous neighborhood in the normal nodes \cite{qiao2024truncated}. By contrast, the presence of anomalous connections and/or attributes makes abnormal nodes deviate significantly from their neighbors. Therefore, for a target node, its latent attributes (\ie, node embedding) are more predictable based on the latent attributes of its neighbors if the node is a normal node, compared to abnormal nodes. The neighborhood-based latent attribute prediction is thus used to measure the normality for GAD. As shown in our experiments (see Figures \ref{motivation}(b)-(d) and Tables \ref{gen_auroc}), it is a generalized anomaly measure that works effectively across graphs. However, due to the existence of irrelevant and noisy attribute information in the original attribute space, the attribute prediction is not as effective as expected in the simply projected space. To address this issue, we propose to learn discriminative prompts via the latent attribute prediction task to enhance the effectiveness of this anomaly measure. 

To achieve this, we first design a simple GNN $g$, a neighborhood aggregation network, to generate the aggregated neighborhood embedding of each target node. Specifically, given a graph $\mathcal{G} = (A, \bar{X})$, the aggregated neighborhood embeddings for each node are obtained as follows:
\begin{equation}\label{neighbor}
\tilde{Z} = g(\mathcal{G})= \tilde{A}\bar{X}W\, ,
\end{equation}
where $\tilde{Z}$ is the aggregated representation of neighbors, $\tilde{A} = (D)^{-1}A$ is the normalized adjacency matrix with $D$ being a diagonal matrix and its elements $D_{kk} = \sum_{j} A_{kj}$, and $W$ is the learnable parameters. Compared to conventional GNNs such as GCN \cite{kipf2016semi} and SGC \cite{wu2019simplifying}, we do not require $\tilde{A}$ to be self-looped and symmetrically normalized as we aim to obtain the aggregated representation of all the neighbors for each node. To design the latent node attribute prediction task, we further obtain the latent attributes of each node as follows:
\begin{equation}\label{mlp}
Z = \bar{X}W\, ,
\end{equation}
where $Z$ serves as the prediction ground truth for the latent attribute prediction task. The adjacency matrix $A$ is discarded to avoid carrying neighborhood-based attribute information into $Z$ which would lead to ground truth leakage in this prediction task. We further propose to utilize the cosine similarity to measure this neighborhood-based latent attribute predictability for each node:
\begin{equation}\label{graphcompatibility}
s_i = \text{sim}(\mathbf{z}_i, \tilde{\mathbf{z}}_i) = \frac{\mathbf{z}_i(\tilde{\mathbf{z}}_i)^T}{\|\mathbf{z}_i\|\|\tilde{\mathbf{z}}_i\|}\, ,
\end{equation}
where $\mathbf{z}$ and $\tilde{\mathbf{z}}_i$ are the $i$-th node embeddings in $Z$ and $\tilde{Z}$ respectively. A higher similarity denotes that the target node can be well predicted by its neighbors and indicates the target is normal with a higher probability. Therefore, we directly utilize the similarity to measure the normal score of the nodes. 

\paragraph{GNN Pre-training.}
To build generalist models, we pre-train the above neighborhood aggregation network via graph contrastive learning due to the ability to obtain robust and transferable models \cite{you2020graph,zhu2020deep} across graphs. Without pre-training, the dataset-specific knowledge would be captured by the model as it is directly optimized based on the neighborhood-based latent attribute prediction of normal and abnormal nodes, limiting the generalizability of the model to other graphs.

\paragraph{Neighborhood Prompting via Latent Attribute Prediction.}
After the pre-training, we aim to further learn more generalized normal and abnormal patterns via prompt tuning. Thus, we devise learnable prompts appended to the attributes of the neighboring nodes of the target nodes, namely \textit{neighborhood prompts}, for learning robust and discriminative patterns that can detect anomalous nodes in different unseen graphs without any re-training during inference.

Specifically, neighborhood prompting aims to learn some prompt tokens that help maximize the neighborhood-based latent prediction of normal nodes while minimizing that of abnormal nodes simultaneously. To this end, the prompt is designed as a set of shared and learnable tokens that can be incorporated into the normalized node attributes. Formally, the neighborhood prompts are represented as $P = [\mathbf{p}_1, \ldots, \mathbf{p}_k]^T \in \mathbb{R}^{K\times d^{'}}$ where $K$ is the number of vector-based tokens $\mathbf{p}_i$. For each node in $\mathcal{G} = (A, \bar{X})$, the node attributes in the unified feature space are augmented by the weighted combination of these tokens, with the weights obtained from $K$ learnable linear projections:
\begin{equation}\label{addprompt}
    \hat{\mathbf{x}}_i = \bar{\mathbf{x}}_i + \sum_{j}^K \alpha_j \mathbf{p}_j\, , \ \ \alpha_j = \frac{e^{(\mathbf{w}_j)^T\mathbf{x}_i^t}}{\sum_{l}^K e^{(\mathbf{w}_l)^T\mathbf{x}_i^t}}\, ,
\end{equation}
where $\alpha_j$ denotes the importance score of the token $\mathbf{p}_j$ in the prompt and $\mathbf{w}_j$ is a learnable projection. For convenience, we denote the graph modified by the graph prompt as $\tilde{\mathcal{G}} = (A, \bar{X}+P)$. Then, $\tilde{\mathcal{G}}$ is fed into the frozen pre-trained model $g$ to obtain the corresponding aggregated embeddings $\tilde{Z}$ and node latent attributes $Z$ via Eq.(\ref{neighbor}) and Eq.(\ref{mlp}) respectively to measure the attribute predictability. To further enhance the representation discrimination, a transformation layer $h$ is applied on the learned $\tilde{Z}$ and $Z$ to transform them into a more anomaly-discriminative feature space,
\begin{equation}\label{transform}
    \tilde{Z} = h(\tilde{Z})\, , \ \ \ \ Z = h(Z)\, .
\end{equation}
The transformed representations are then used to measure the latent node attribute predictability with Eq.(\ref{graphcompatibility}). To optimize $P$ and $h$, we employ the following training objective,
\begin{equation}\label{ggad_obj}
    \min_{P,h} \ \ \sum \ell(\mathbf{z}_i, \tilde{\mathbf{z}}_i)\, ,
\end{equation}
where $\ell(\mathbf{z}_i, \tilde{\mathbf{z}}_i) = -\text{sim}(\mathbf{z}_i, \tilde{\mathbf{z}}_i)$ if $\mathbf{y}_i = 0$, and $\ell(\mathbf{z}_i, \tilde{\mathbf{z}}_i) = \text{sim}(\mathbf{z}_i,\tilde{\mathbf{z}}_i)$ if $\mathbf{y}_i = 1$.

\subsection{Training and Inference of UNPrompt}
\noindent\textbf{Training.} Given $\mathcal{G}_{\text{train}}$, a neighborhood aggregation network $g$ is optimized via graph contrastive learning. Then, the neighborhood prompts $P$ and the transformation layer $h$ are optimized to capture the graph-agnostic normal and abnormal patterns while keeping the pre-trained model $g$ frozen. In this way, the transferable knowledge of the pre-trained $g$ is maintained, while the neighborhood prompt learning helps learn the generalized normal and abnormal patterns.

\noindent\textbf{Inference.} 
During inference, given $\mathcal{G}_{\text{test}}^{(i)}\in\mathcal{T}_{\text{test}}$, the node attributes are first aligned.
Then, the test graph $\mathcal{G}_{\text{test}}^{(i)}$ is augmented with the learned neighborhood prompt $P$ 
and fed into the model $g$ and the transformation layer $h$ to obtain the neighborhood aggregated representations and the latent node attributes. Finally, the similarity (Eq.(\ref{graphcompatibility})) is used as the normal score for anomaly detection. Note that the inference does not require any further re-training and labeled nodes of $\mathcal{G}_{\text{test}}^{(i)}$. 

\section{Time Complexity Analysis}
UNPrompt consists of GNN pre-training, neighborhood prompt learning, and the transformation layer. The overall time complexity is $\mathcal{O}(4E_1(|A|d_h+Nd_hd^{'}+6Nd_h^2) + 2E_2(|A|d_h+Nd_hd^{'} + 2KNd^{'}+Nd_h^2))$, where $d_h$ is the number of hidden units, $|A|$ returns of the number of edges of the $\mathcal{G}_{\text{train}}$, $N$ is the number of nodes, $d^{'}$ represents the predefined dimensionality of node attributes, $E_1$ is the number of pre-training epoch, $K$ is the size of prompt and $E_2$ is the training epochs for prompt learning. Theoretical and empirical results on computational cost are presented in the appendix.

\begin{table*}[!ht]
\begin{center}
\resizebox{0.9\textwidth}{!}{
\begin{tabular}{c|l|ccccccc|c}
\hline
\hline
\multirow{2}*{\textbf{Metric}}&\multirow{2}*{\textbf{Method}} & \multicolumn{7}{c|}{\textbf{Dataset}}\\
&&Amazon &Reddit &Weibo & YelpChi &Aamzon-all &YelpChi-all&Disney&Avg.\\
\hline
\multirow{13}{*}{AUROC}
&\multicolumn{9}{c}{\cellcolor[HTML]{EFEFEF}Unsupervised Methods}\\
&AnomalyDAE (ICASSP'20) &0.5818$_{\pm 0.039}$&0.5016$_{\pm 0.032}$&\underline{0.7785}$_{\pm 0.058}$&0.4837$_{\pm 0.094}$&0.7228$_{\pm 0.023}$&0.5002$_{\pm 0.018}$&0.4853$_{\pm 0.003}$&\underline{0.5791}\\
&CoLA (TNNLS'21)   &0.4580$_{\pm 0.054}$&0.4623$_{\pm 0.005}$&0.3924$_{\pm 0.041}$&0.4907$_{\pm 0.017}$&0.4091$_{\pm 0.052}$&0.4879$_{\pm 0.010}$&0.4696$_{\pm 0.065}$&0.4529\\
&HCM-A (ECML-PKDD'22)  &0.4784$_{\pm 0.005}$&0.5387$_{\pm 0.041}$&0.5782$_{\pm 0.048}$&0.5000$_{\pm 0.000}$&0.5056$_{\pm 0.059}$&0.5023$_{\pm 0.005}$&0.2014$_{\pm 0.015}$&0.4721\\
&TAM (NeurIPS'23)  &0.4720$_{\pm 0.005}$&\textbf{0.5725}$_{\pm 0.004}$&0.4867$_{\pm 0.028}$&0.5035$_{\pm 0.014}$&0.7543$_{\pm 0.002}$&0.4216$_{\pm 0.002}$&0.4773$_{\pm 0.003}$&0.5268\\
&{GADAM} (ICLR'24)&{\underline{0.6646}$_{\pm 0.063}$}&{0.4532$_{\pm 0.024}$}&{0.3652$_{\pm 0.052}$}&{0.3376$_{\pm 0.012}$}&{0.5959$_{\pm 0.080}$}&{0.4829$_{\pm 0.016}$}&0.4288$_{\pm 0.023}$&0.4755\\
&\multicolumn{9}{c}{\cellcolor[HTML]{EFEFEF}Supervised Methods}\\
&GCN (ICLR'17)   &0.5988$_{\pm 0.016}$&\underline{0.5645}$_{\pm 0.000}$&0.2232$_{\pm 0.074}$&0.5366$_{\pm 0.019}$&0.7195$_{\pm 0.002}$&\underline{0.5486}$_{\pm 0.001}$&0.5000$_{\pm 0.000}$&0.5273\\
&GAT (ICLR'18) &0.4981$_{\pm 0.008}$&0.5000$_{\pm 0.025}$&0.4521$_{\pm 0.101}$&\underline{0.5871}$_{\pm 0.016}$&0.5005$_{\pm 0.012}$&0.4802$_{\pm 0.004}$&0.5175$_{\pm 0.054}$&0.5051\\
&BWGNN (ICML'22) &0.4769$_{\pm 0.020}$&0.5208$_{\pm 0.016}$&0.4815$_{\pm 0.108}$&0.5538$_{\pm 0.027}$&0.3648$_{\pm 0.050}$&0.5282$_{\pm 0.015}$&0.6073$_{\pm 0.026}$&0.5048\\
&GHRN (WebConf'23)  &0.4560$_{\pm 0.033}$&0.5253$_{\pm 0.006}$&0.5318$_{\pm 0.038}$&0.5524$_{\pm 0.020}$&0.3382$_{\pm 0.085}$&0.5125$_{\pm 0.016}$&0.5336$_{\pm 0.030}$&0.4928\\
&{XGBGraph} (NeurIPS'23)&{0.4179$_{\pm 0.000}$}&{0.4601$_{\pm 0.000}$}&{0.5373$_{\pm 0.000}$}&{0.5722$_{\pm 0.000}$}&{\underline{0.7950}$_{\pm 0.000}$}&{0.4945$_{\pm 0.000}$}&\textbf{0.6692}$_{\pm 0.000}$&0.5637\\
&GraphPrompt (WebConf'23) &0.4904$_{\pm 0.001}$&0.4677$_{\pm 0.001}$&0.6135$_{\pm 0.008}$&0.3935$_{\pm 0.002}$&0.3215$_{\pm 0.001}$&0.4976$_{\pm 0.000}$&0.5192$_{\pm 0.002}$&0.4719\\
\cline{2-10}
&UNPrompt (Ours)    &\textbf{0.7525}$_{\pm 0.016}$&0.5337$_{\pm 0.002}$&\textbf{0.8860}$_{\pm 0.007}$&\textbf{0.5875}$_{\pm 0.016}$&\textbf{0.7962}$_{\pm 0.022}$&\textbf{0.5558}$_{\pm 0.012}$&\underline{0.6412}$_{\pm 0.030}$&\textbf{0.6790}\\
\hline
\hline
\multirow{13}{*}{AUPRC}
&\multicolumn{9}{c}{\cellcolor[HTML]{EFEFEF}Unsupervised Methods}\\
&AnomalyDAE (ICASSP'20) &0.0833$_{\pm 0.015}$&0.0327$_{\pm 0.004}$&\underline{0.6064}$_{\pm 0.031}$&0.0624$_{\pm 0.017}$&0.1921$_{\pm 0.026}$&0.1484$_{\pm 0.009}$&0.0566$_{\pm 0.000}$&\underline{0.1688}\\
&CoLA (TNNLS'21)   &0.0669$_{\pm 0.002}$&0.0391$_{\pm 0.004}$&0.1189$_{\pm 0.014}$&0.0511$_{\pm 0.000}$&0.0861$_{\pm 0.019}$&0.1466$_{\pm 0.003}$&0.0701$_{\pm 0.023}$&0.0827\\
&HCM-A (ECML-PKDD'22) &0.0669$_{\pm 0.002}$&0.0391$_{\pm 0.004}$&0.1189$_{\pm 0.014}$&0.0511$_{\pm 0.000}$&0.0861$_{\pm 0.019}$&0.1466$_{\pm 0.003}$&0.0355$_{\pm 0.001}$&0.0777\\
&TAM (NeurIPS'23) &0.0666$_{\pm 0.001}$&\underline{0.0413}$_{\pm 0.001}$&0.1240$_{\pm 0.014}$&0.0524$_{\pm 0.002}$&{0.1736}$_{\pm 0.004}$&0.1240$_{\pm 0.001}$&0.0628$_{\pm 0.001}$&0.0921\\
&GADAM (ICLR'24) &{0.1562$_{\pm 0.103}$}&{0.0293$_{\pm 0.001}$}&{0.0830$_{\pm 0.005}$}&{0.0352$_{\pm 0.001}$}&{0.1595$_{\pm 0.121}$}&{0.1371$_{\pm 0.006}$}&0.0651$_{\pm 0.012}$&0.0951\\
&\multicolumn{9}{c}{\cellcolor[HTML]{EFEFEF}Supervised Methods}\\
&GCN (ICLR'17) &\underline{0.0891}$_{\pm 0.007}$&\textbf{0.0439}$_{\pm 0.000}$&0.1109$_{\pm 0.020}$&0.0648$_{\pm 0.009}$&0.1536$_{\pm 0.002}$&\underline{0.1735}$_{\pm 0.000}$&0.0484$_{\pm 0.000}$&0.0977\\
&GAT (ICLR'18) &0.0688$_{\pm 0.002}$&0.0329$_{\pm 0.002}$&0.1009$_{\pm 0.017}$&\textbf{0.0810}$_{\pm 0.005}$&0.0696$_{\pm 0.001}$&0.1400$_{\pm 0.002}$&0.0530$_{\pm 0.004}$&0.0780\\
&BWGNN (ICML'22) &0.0652$_{\pm 0.002}$&0.0389$_{\pm 0.003}$&0.2241$_{\pm 0.046}$&0.0708$_{\pm 0.018}$&0.0586$_{\pm 0.003}$&0.1605$_{\pm 0.005}$&0.0624$_{\pm 0.003}$&0.0972\\
&GHRN (WebConf'23) &0.0633$_{\pm 0.003}$&0.0407$_{\pm 0.002}$&0.1965$_{\pm 0.059}$&0.0661$_{\pm 0.010}$&0.0569$_{\pm 0.006}$&0.1505$_{\pm 0.005}$&0.0519$_{\pm 0.003}$&0.0894\\
&{XGBGraph} (NeurIPS'23) &{0.0536$_{\pm 0.000}$}&{0.0330$_{\pm 0.000}$}&{0.2256$_{\pm 0.000}$}&{0.0655$_{\pm 0.000}$}&{\underline{0.2307}$_{\pm 0.000}$}&{0.1449$_{\pm 0.000}$}&\underline{0.1215}$_{\pm 0.000}$&0.1250\\
&GraphPrompt (WebConf'23) &0.0661$_{\pm 0.000}$&0.0334$_{\pm 0.000}$&0.2014$_{\pm 0.004}$&0.0382$_{\pm 0.000}$&0.0666$_{\pm 0.000}$&0.1542$_{\pm 0.000}$&0.0617$_{\pm 0.001}$&0.0888\\
\cline{2-10}
&UNPrompt (Ours)     &\textbf{0.1602}$_{\pm 0.013}$&0.0351$_{\pm 0.000}$&\textbf{0.6406}$_{\pm 0.026}$&\underline{0.0712}$_{\pm 0.008}$&\textbf{0.2430}$_{\pm 0.028}$&\textbf{0.1810}$_{\pm 0.012}$&\textbf{0.1236}$_{\pm 0.031}$&\textbf{0.2078}\\
\hline 
\hline
\end{tabular}
}
\end{center}
\caption{AUROC and AUPRC results on seven real-world GAD datasets with the models trained on Facebook only. For each dataset, the best performance per column within each metric is boldfaced, with the second-best underlined. ``Avg'' denotes the average performance.}
\label{gen_auroc}
\end{table*}

\section{Experiments}

\subsection{Performance on Zero-shot Generalist GAD}
\paragraph{Datasets.}
We evaluate UNPrompt on several real-world GAD datasets from diverse social networks, online shopping co-review networks, and co-purchase networks. Specifically, the social networks include Facebook \cite{xu2022contrastive}, Reddit \cite{kumar2019predicting} and Weibo \cite{kumar2019predicting}. The co-review networks consist of Amazon \cite{mcauley2013amateurs}, YelpChi \cite{rayana2015collective}, Amazon-all and YelpChi-all. Disney \cite{sanchez2013statistical} is a co-purchase network.

\paragraph{Competing Methods.}
Since there is no zero-shot generalist GAD method, a set of state-of-the-art (SotA) unsupervised and supervised competing methods are employed for comparison. The unsupervised methods comprise reconstruction-based AnomalyDAE \cite{fan2020anomalydae}, contrastive learning-based CoLA \cite{liu2021anomaly}, hop prediction-based HCM-A \cite{huang2022hop}, local affinity-based TAM \cite{qiao2024truncated} and GADAM \cite{chen2024boosting}. Supervised methods include two conventional GNNs -- GCN \cite{kipf2016semi} and GAT \cite{velivckovic2017graph} -- and three SotA GAD GNNs -- BWGNN \cite{tang2022rethinking}, GHRN \cite{gao2023addressing} and XGBGraph \cite{tang2023gadbench}. As a graph prompting method, GraphPrompt \cite{liu2023graphprompt} is also used. Following \cite{liu2024arc,qiao2024truncated,qiao2024deep}, two widely-used metrics, AUROC and AUPRC, are used to evaluate the performance of all methods. For both metrics, the higher value denotes the better performance. Moreover, for each method, we report the average performance with standard deviations after 5 independent runs with different random seeds.

\paragraph{Implementation Details.}
For a fair comparison, the common dimensionality is set to eight for all methods, and SVD is used for feature projection. The number of GNN layers is set to one and the number of hidden units is 128. The transformation layer is implemented as a one-layer MLP with the same number of hidden units. The size of the neighborhood prompt is set to one. Results for other hyperparameter settings are presented in the appendix. For all baselines, their recommended optimization and hyperparameter settings are used. Note that GraphPrompt is a graph prompting method. To adapt it for generalist GAD, after pre-training, we further perform the prompt learning in the source graph for supervised GAD. Then, the pre-trained model and learned prompts are used for generalist anomaly detection. UNPrompt and all competing methods are trained on Facebook and then directly tested on the other GAD datasets. Facebook is used in the training since its anomaly patterns are more generic.

\paragraph{Main Results.}
The results of all methods are presented in Table~\ref{gen_auroc} and we can have the following observations. (1) Under the proposed generalist GAD scenario where a model is trained on a single dataset and evaluated on seven other datasets, all the competing baselines fail to work well, demonstrating that it is very challenging to build a generalist GAD model that generalizes across different datasets under the zero-shot setting. (2) For supervised methods, the simple GCN achieves better performance than the specially designed GAD GNNs. This can be attributed to more dataset-specific knowledge being captured in these specialized GAD models, limiting their generalization capacity to the unseen testing graphs. (3) Unsupervised methods perform more stably than supervised methods across the target graphs. This is because the unsupervised objectives are closer to the shared anomaly patterns across graphs compared to the supervised ones, especially for TAM which employs a fairly generalized local affinity-based objective for GAD. 
\begin{table}[!ht]
\begin{center}
\resizebox{0.45\textwidth}{!}{
\begin{tabular}{l|cccc|c}
\toprule
\hline
Method &Amazon &Reddit &Weibo & YelpChi &Avg.\\
\hline
UNPrompt    &\textbf{0.7525}&0.5337&\textbf{0.8860}&\textbf{0.5875}&\textbf{0.6899}\\
\hline
\ \ w/o Normalization  &0.4684&0.5006&0.1889&0.5620&0.4300\\
\ \ w/o Pre-training  &0.5400&0.5233&0.5658&0.4672&0.5241\\
\ \ w/o Prompt        &0.5328&0.5500&0.4000&0.4520&0.4837\\
\ \ w/o Transformation &0.7331&\textbf{0.5556}&0.7406&0.5712&0.6501\\
\hline
\bottomrule
\end{tabular}}
\end{center}
\caption{AUROC results of UNPrompt and its four variants.}
\label{albations}
\end{table}
(4) Despite being a graph prompting method, GraphPrompt fails to achieve promising performance for zero-shot generalist GAD. This is because no label information is provided and no further model optimization is performed during the inference. (5) The proposed UNPrompt demonstrates strong and stable generalist GAD capacity across graphs from different distributions and domains. Specifically, UNPrompt achieves the best AUROC performance on 5 out of 7 datasets, and the average performance outperforms the best-competing method by over 9\%. The superiority is attributed to i) the proposed coordinate-wise normalization effectively aligns the features across graphs, and ii) the shared generalized normal and abnormal patterns are well captured in the neighborhood prompts.

\paragraph{Ablation Study.}
To evaluate the importance of each component in UNPrompt, we design four variants, \ie, w/o coordinate-wise normalization, w/o graph contrastive learning-based pre-training, w/o neighborhood prompts, and w/o transformation layer. The results of these variants are reported in the Table~\ref{albations}. Due to the page limits, four datasets are used, including Amazon, Reddit, Weibo and YelpChi. From the table, we can see that all four components contribute to the overall superior performance of UNPrompt. More specifically, (1) without the coordinate-wise normalization, the method fails to calibrate the distributions of diverse node attributes into a common space, leading to large performance drop across all datasets. (2) Besides the semantics alignment, the graph contrastive learning-based pre-training ensures our GNN network is transferable to other graphs instead of overfitting to the training graph. As expected, the performance of the variant without pre-training also drops significantly. (3) If the neighborhood prompts are removed, the learning of latent node attribute prediction is ineffective for capturing generalized normal and abnormal patterns. (4) The variant without the transformation layer achieves inferior performance on nearly all the datasets, demonstrating the importance of mapping the features into a more anomaly-discriminative space.

\begin{figure}
\centering
\subfigure[]{\label{sensitivity}\includegraphics[width=0.23\textwidth]{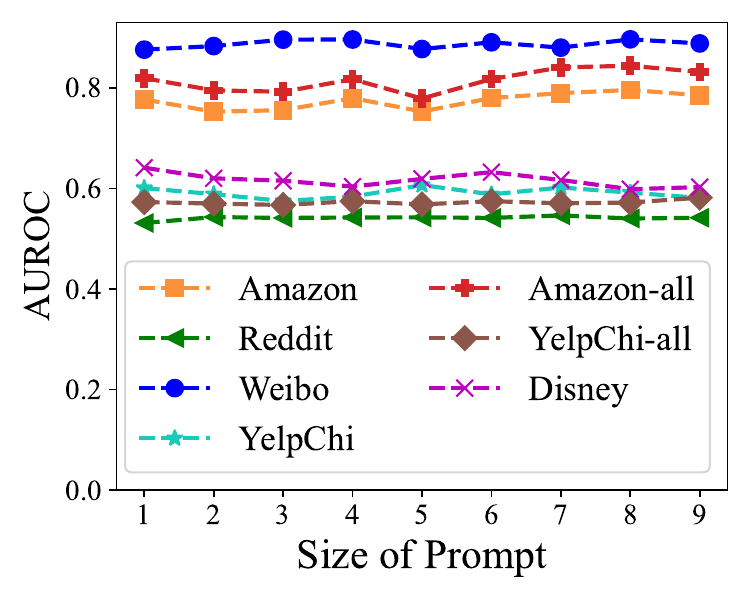}} 
\subfigure[]{\label{diff_obj}\includegraphics[width=0.23\textwidth]{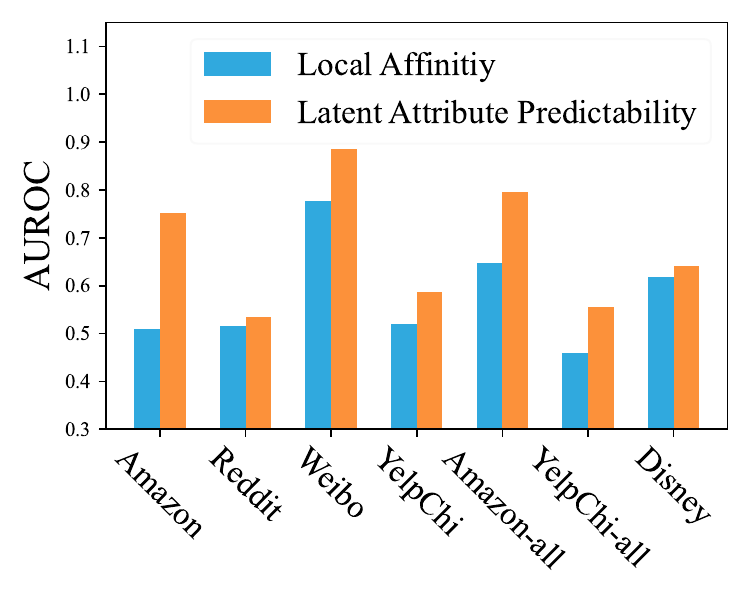}}  
\caption{(\textbf{a}) AUROC results of UNPrompt w.r.t. varying neighborhood prompt size. (\textbf{b}). The AUROC performance of generalist GAD with different prompt learning objectives.} 
\end{figure}

\paragraph{Sensitivity w.r.t the Neighborhood Prompt Size.} 
We evaluate the sensitivity of UNPrompt w.r.t the size of the neighborhood prompts \ie, $K$. We vary $K$ in the range of $[1, 9]$ and report the results in Figure~\ref{sensitivity}. It is clear that the performances on Reddit, Weibo and YelpChi-all remain stable with varying sizes of neighborhood prompts while the other datasets show slight fluctuation, demonstrating that the generalized normal and abnormal patterns can be effectively captured in our neighborhood prompts even with a small size. 

\paragraph{Prompt learning using latent attribute prediction vs. alternative graph anomaly measures.}
To further justify the effectiveness of latent attribute predictability on learning generalized GAD patterns in our prompt learning module, we compare our learnable anomaly measure to the recently proposed anomaly measure, local node affinity \cite{qiao2024truncated}. All modules of UNPrompt are fixed, with only the latent attribute prediction task replaced as the maximization of local affinity as in TAM. The results are presented in Figure~\ref{diff_obj}. We can see that the latent attribute predictability consistently and significantly outperforms the local affinity-based measure across all graphs, demonstrating its superiority in learning generalized patterns for generalist GAD. 

\subsection{Performance in Unsupervised GAD}
UNPrompt can also be easily adapted to conventional unsupervised GAD. Experimental details are provided in the appendix. The results of comparing UNPrompt to SotA unsupervised methods are presented in Table~\ref{unsupervised}. Despite being a generalist method, UNPrompt also works very well as a specialized GAD model. UNPrompt substantially outperforms all the baselines on all datasets. Particularly, the average performance of UNPrompt surpasses the best-competing method by over 2\%, showing that the latent node attribute predictability is a generalized GAD measure that also holds for different graphs under unsupervised settings.

\begin{table}[t]
\begin{center}
\resizebox{0.47\textwidth}{!}{
\begin{tabular}{c|cccccc|c}
\toprule
\hline
\multirow{2}*{\textbf{Method}} & \multicolumn{6}{c|}{\textbf{Dataset}}\\
&Amazon &Facebook&Reddit & YelpChi &Amazon-all & YelpChi-all & Avg. \\
\hline
iForest  & 0.5621 &0.5382 &0.4363 &0.4120 & 0.1914  & 0.3617 &0.4169  \\
ANOMALOUS &0.4457 &0.9021 &0.5387 &0.4956 &0.3230 & 0.3474 &0.5087  \\
DOMINANT  &0.5996 &0.5677 &0.5555  &0.4133 & 0.6937   &0.5390  &0.5615  \\
CoLA   &0.5898 &0.8434 &\underline{0.6028} &0.4636 & 0.2614    &0.4801 &0.5402 \\
SL-GAD &0.5937 & 0.7936 & 0.5677 & 0.3312 & 0.2728  &0.5551 &0.5190  \\
HCM-A  &0.3956 &0.7387 &0.4593 &0.4593 & 0.4191 & 0.5691 &0.5069  \\
ComGA  &0.5895 & 0.6055 & 0.5453 & 0.4391 & 0.7154  &0.5352 &  0.5716\\
TAM    &\underline{0.7064} &\underline{0.9144} &0.6023   &\underline{0.5643} &\underline{0.8476}  &\underline{0.5818} &\underline{0.7028} \\
\hline
UNPrompt (Ours) &\textbf{0.7335}&\textbf{0.9379}&\textbf{0.6067}&\textbf{0.6223} &\textbf{0.8516} &\textbf{0.6084} &\textbf{0.7267}\\
\hline
\bottomrule
\end{tabular}}
\end{center}
\caption{AUROC results of unsupervised GAD methods.}
\label{unsupervised}
\end{table}

\section{Conclusion}
In this paper, we propose a novel zero-shot generalist GAD method that trains one detector on a single dataset and can effectively generalize to other unseen target graphs without any further re-training or labeled nodes of target graphs during inference. The attribute inconsistency and the absence of generalized anomaly patterns are the main obstacles for generalist GAD. To address these issues, two main modules are proposed, \ie, coordinate-wise normalization-based attribute unification and neighborhood prompt learning. The first module aligns node attribute dimensionality and semantics, while the second module captures generalized normal and abnormal patterns via the neighborhood-based latent node attribute prediction. Extensive experiments on both generalist and unsupervised GAD demonstrate the effectiveness of UNPrompt.

\section*{Acknowledgments}
This research is supported by ARC under Grant DP240101349, the Ministry of Education, Singapore under its Tier-1 Academic Research Fund (24-SIS-SMU-008), A*STAR under its MTC YIRG Grant (M24N8c0103), and the Lee Kong Chian Fellowship (T050273).

\bibliographystyle{named}
\bibliography{ijcai25}

\appendix

\section{Graph Similarity}
In addition to the visualization results presented in Figure 1 in the main paper,  we further provide the distributional similarity of various graphs in this section. Specifically, for dimension-aligned graphs across different distributions and domains, we measure their distributional similarity to analyze their diverse semantics.

Given a graph $\mathcal{G}^{(i)} = (A^{(i)}, \tilde{X}^{(i)})$, the coordinate-wise mean $\boldsymbol{\mu}^{(i)} = [{\mu}^{(i)}_1, \ldots, {\mu}^{(i)}_{d^{'}}]$ and variance $\boldsymbol{\sigma}^{(i)} = [{\sigma}^{(i)}_1, \ldots, {\sigma}^{(i)}_{d^{'}}]$ of $\tilde{X}^{(i)}$ are calculated and concatenated to form the distributional vector of $\mathcal{G}^{(i)}$, \ie, $\mathbf{d}_i = [\boldsymbol{\mu}^{(i)}, \boldsymbol{\sigma}^{(i)}]$. Then, the distribution similarity between $\mathcal{G}^{(i)}$ and $\mathcal{G}^{(j)}$ is measured via the cosine similarity,
\begin{equation}
    s_{ij} = \text{sim}(\mathbf{d}_i, \mathbf{d}_j)\, .
\end{equation}

\begin{figure}[ht]
\centering
\subfigure[]{\label{dist_sim}\includegraphics[width=0.23\textwidth]{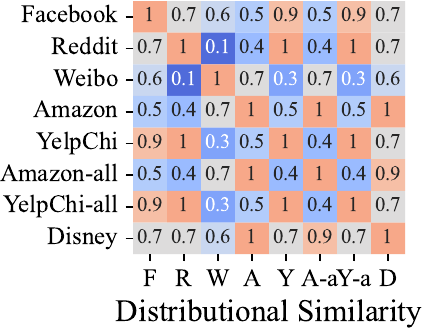}} 
\subfigure[]{\label{dist_sim_nor}\includegraphics[width=0.23\textwidth]{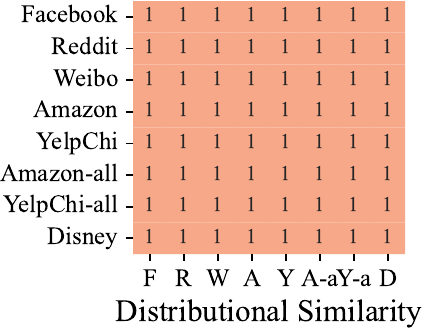}} 
\caption{\textbf{(a)} Distributional similarity between different graphs without coordinate-wise normalization. \textbf{(b)} Distributional similarity between different graphs with the coordinate-wise normalization.}
\end{figure}

The distributional similarities between graphs from different domains or distributions are shown in Figure~\ref{dist_sim}.. From the figure, we can see that the distributional similarities are typically small, demonstrating the diverse semantics of node features across graphs. Noth that, for Amazon \& Amazon-all and YelpChi \& YelpChi-all, their distribution similarity is one, which can be attributed to the fact that they are from the same distributions respectively, but with different numbers of nodes and structures.

To reduce the semantic gap among graphs for generalist GAD, we propose to calibrate the distributions of all graphs into the same frame with coordinate-wise normalization. The distributional similarity with normalization is illustrated in Figure~\ref{dist_sim_nor}. It is clear that the node attributes share the same distribution after the normalization. In this way, the generalist model can better capture the shared GAD patterns and generalize to different target graphs, as demonstrated in our experimental results.

\section{Details on Pre-training of Neighborhood Aggregation Networks}\label{graphcontrastivelearning}

We pre-train the neighborhood aggregation network $g$ via graph contrastive learning \cite{zhu2020deep} for subsequent graph prompt learning so that the generic normality and abnormality can be captured in the prompts. 

Specifically, given the training graph $\mathcal{G} = (A, X)$, to construct contrastive views for graph contrastive learning, two widely used graph augmentations are employed, \ie, edge removal and attribute masking \cite{zhu2020deep}. The edge removal randomly drops a certain portion of existing edges in $\mathcal{G}$ and the attribute masking randomly masks a fraction of dimensions with zeros in node attributes, \ie,
\begin{equation}
    \hat{A} = A \circ R\, , \ \ \  \hat{X} = [\mathbf{x}_1\circ \mathbf{m}, \ldots, \mathbf{x}_N \circ \mathbf{m}]^T\, ,  
\end{equation}
where $R \in \{0, 1\}^{N\times N}$ is the edge masking matrix whose entry is drawn from a Bernoulli distribution controlled by the edge removal probability, $\mathbf{m} \in \{0,1\}^d$ is the attribute masking vector whose entry is independently drawn from a Bernoulli distribution with the attribute masking ratio, and $\circ$ denotes the Hadamard product. 

By applying the graph augmentations to the original graph, the corrupted graph $\hat{\mathcal{G}} = (\hat{A}, \hat{X})$ forms the contrastive view for the original graph $\mathcal{G} = (A, X)$. Then, $\hat{\mathcal{G}}$ and $\mathcal{G}$ are fed to the shared model $g$ followed by the non-linear projection to obtain the corresponding node embeddings, \ie, $\hat{Z}^\prime$ and $Z^\prime$. For graph contrastive learning, the embeddings of the same node in different views are pulled closer while the embeddings of other nodes are pushed apart. The pairwise objective for each node pair $(\hat{\mathbf{z}}^\prime_i, \mathbf{z}^\prime_i)$ can be formulated as:
\begin{align}
&\ell(\hat{\mathbf{z}}^\prime_i, \mathbf{z}^\prime_i) = \nonumber \\
&- \log \frac{e^{\text{sim}(\hat{\mathbf{z}}^\prime_i, \mathbf{z}^\prime_i)/\tau}}{e^{\text{sim}(\hat{\mathbf{z}}^\prime_i, \mathbf{z}^\prime_i)/\tau} + \sum_{j\neq i}^N e^{\text{sim}(\hat{\mathbf{z}}^\prime_i, \mathbf{z}^\prime_j)/\tau} + \sum_{j\neq i}^N e^{\text{sim}(\hat{\mathbf{z}}^\prime_i, \hat{\mathbf{z}}^\prime_j)/\tau}}\, , \end{align}
where $\text{sim}(\cdot)$ represents the cosine similarity and $\tau$ is a temperature hyperparameter. Therefore, the overall objective can be defined as follows:
\begin{equation}\label{contrastive}
    \mathcal{L}_{\text{contrast}} = \frac{1}{2N} \sum_{i=1}^N (\ell(\hat{\mathbf{z}}^\prime_i, \mathbf{z}^\prime_i) + \ell(\mathbf{z}^\prime_i, \hat{\mathbf{z}}^\prime_i))\, .
\end{equation}
With the objective Eq.(\ref{contrastive}), the model $g$ is optimized to learn transferable discriminative representations of nodes.

\section{Algorithms}\label{algo}
The training and inference processes of UNPrompt are summarized in Algorithms \ref{alg} and Algorithm \ref{alg2}, respectively.

\begin{algorithm}[ht]
\small
\caption{Training of UNPrompt}
\begin{algorithmic}[1]
\label{alg}
\STATE {\textbf{Input:}} Training graph $\mathcal{G}_{\text{train}}=(A, X)$; training epoch $E$\\
\STATE {\textbf{Output:}} Neighborhood aggregation network $g$, graph prompts $P= [\mathbf{p}_1, \ldots, \mathbf{p}_K]$, and transformation $h$.
\STATE Perform feature unification of $X$.
\STATE Pre-train $g$ on $\mathcal{G}_{\text{train}}$ with graph contrastive learning in Eq.(~\ref{contrastive}).
\STATE Keep model $g$ frozen.
\FOR{$e=1,\ldots, E$}
\STATE Obtain modified node attribute with prompts via Eq.(6).\\
\STATE Obtain the neighborhood aggregated representation $\tilde{Z}$ via Eq.(3).\\
\STATE Obtain the node representations $Z$ via Eq.(4).\\
\STATE Transform $\tilde{Z}$ and $Z$ with $h$ via Eq.(7).
\STATE Optimize $P$ and $h$ by minimizing Eq.(8).
\ENDFOR
\end{algorithmic}
\end{algorithm}

\begin{algorithm}[ht]
\small
\caption{Inference of UNPrompt}
\begin{algorithmic}[1]
\label{alg2}
\STATE {\textbf{Input:}} Testing graphs $\mathcal{T}_{\text{test}} = \{\mathcal{G}_{\text{test}}^{(1)}, \ldots, {\mathcal{G}_{\text{test}}^{(n)}}\}$, neighborhood aggregation network $g$, graph prompts $P= [\mathbf{p}_1, \ldots, \mathbf{p}_K]$, and transformation $h$.
\STATE {\textbf{Output:}} Normal score of testing nodes.
\FOR{$\mathcal{G}_{\text{test}}^{(i)} = (A^{(i)}, X^{(i)}) \in \mathcal{T}_{\text{test}}$}
\STATE Perform feature unification of $X^{(i)}$.\\
\STATE Obtain modified node attribute with prompts via Eq.(6).\\
\STATE Obtain the neighborhood aggregated representation $\tilde{Z}^{(i)}$ via Eq.(3).\\
\STATE Obtain the node representations $Z^{(i)}$ via Eq.(4).\\
\STATE Transform $\tilde{Z}^{(i)}$ and $Z^{(i)}$ with $h$ via Eq.(7). \\
\STATE Obtain the normal score of nodes via Eq.(5).
\ENDFOR
\end{algorithmic}
\end{algorithm}

\section{Unsupervised GAD with UNPrompt}\label{unsupervised_gad}
To demonstrate the wide applicability of the proposed method UNPrompt, we further perform unsupervised GAD with UNPrompt which focuses on detecting anomalous nodes within one graph and does not have access to any node labels during training. Specifically, we adopt the same pipeline in the generalist GAD setting, \ie, graph contrastive pertaining and neighborhood prompt learning. Since we focus on anomaly detection on each graph separately, the node attribute unification is discarded for unsupervised GAD. However, the absence of node labels poses a challenge to learning meaningful neighborhood prompts for anomaly detection. To overcome this issue, we propose to utilize the pseudo-labeling technique to guide the prompt learning. Specifically, the normal score of each node is calculated by the neighborhood-based latent attribute predictability after the graph contrastive learning process:
\begin{equation}\label{un_score}
    s_i = \text{sim}(\mathbf{z}_i, \tilde{\mathbf{z}}_i)\, ,
\end{equation}
where $\mathbf{z}_i$ is the node representation learned by graph contrastive learning and $\tilde{\mathbf{z}}_i$ is the corresponding aggregated neighborhood representation. Higher $s_i$ of node $v_i$ typically indicates a higher probability of $v_i$ being a normal node. Therefore, more emphasis should be put on high-score nodes when learning neighborhood prompts. To achieve this, the normal score $s_i$ is transformed into the loss weight $w_i = \text{Sigmoid}(\alpha(s_i - t))$ where $t$ is a threshold and $\alpha$ is the scaling parameter. In this way, $w_i$ would approach 1 if $s_i > t$ and 0 otherwise. Overall, the objective for unsupervised GAD using UNPrompt can be formulated as follows:
\begin{equation}\label{ugad_obj}
\mathcal{L} = \sum_i^N (-w_i\text{sim}(\mathbf{z}_i, \tilde{\mathbf{z}}_i) + \lambda \sum_{j, j\neq i}^N\text{sim}(\mathbf{z}_i, \tilde{\mathbf{z}}_j))\, ,
\end{equation}
where the second term is a regularization term employed to prevent the node embeddings from being collapsed into the same and $\lambda$ is a trade-off hyperparameter. 

Note that we only focus on maximizing the latent attribute predictability of high-score nodes without minimizing the predictability of low-score nodes in the above objective. These low-score nodes could also be normal nodes with high probability as the score from Eq.(\ref{un_score}) is only obtained from the pre-trained model, resulting in the score not being fully reliable. If the predictability is also minimized for these nodes, conflicts would be induced for neighborhood prompt learning, limiting the performance of unsupervised GAD. After optimization, the latent attribute predictability is also directly used as the normal score for the final unsupervised GAD.

\begin{table*}[!ht]
\begin{center}
\resizebox{0.9\textwidth}{!}{
\begin{tabular}{c|l|cccccc|c}
\toprule
\hline
\multirow{2}*{\textbf{Metric}}&\multirow{2}*{\textbf{Method}} & \multicolumn{6}{c|}{\textbf{Dataset}}\\
& &Amazon &Facebook&Reddit & YelpChi &Amazon-all & YelpChi-all & Avg. \\
\hline
\multirow{9}*{AUROC}
&iForest (TKDD'12) & 0.5621$_{\pm 0.008}$ &0.5382$_{\pm 0.015}$ &0.4363$_{\pm 0.020}$ &0.4120$_{\pm 0.040}$ & 0.1914$_{\pm 0.002}$  & 0.3617$_{\pm 0.001}$ &0.4169  \\
&ANOMALOUS (IJCAI'18) &0.4457$_{\pm 0.003}$ &0.9021$_{\pm 0.005}$ &0.5387$_{\pm 0.012}$ &0.4956$_{\pm 0.003}$    &0.3230$_{\pm 0.021}$ & 0.3474$_{\pm 0.018}$ &0.5087  \\
&DOMINANT (SIAM'19) &0.5996$_{\pm 0.004}$ &0.5677$_{\pm 0.002}$ &0.5555$_{\pm 0.011}$  &0.4133$_{\pm 0.010}$  & 0.6937$_{\pm 0.028}$   &0.5390$_{\pm 0.014}$  &0.5615  \\
&CoLA (TNNLS'21)  &0.5898$_{\pm 0.008}$ &0.8434$_{\pm 0.011}$ &\underline{0.6028}$_{\pm 0.007}$ &0.4636$_{\pm 0.001}$ & 0.2614$_{\pm 0.021}$    &0.4801$_{\pm 0.016}$ &0.5402 \\
&SL-GAD (TKDE'21) &0.5937$_{\pm 0.011}$ & 0.7936$_{\pm 0.005}$ & 0.5677$_{\pm 0.005}$ & 0.3312$_{\pm 0.035}$  & 0.2728$_{\pm 0.012}$  &0.5551$_{\pm 0.015}$ &0.5190  \\
&HCM-A (ECML-PKDD'22) &0.3956$_{\pm 0.014}$ &0.7387$_{\pm 0.032}$ &0.4593$_{\pm 0.011}$  &0.4593$_{\pm 0.005}$ & 0.4191$_{\pm 0.011}$ & 0.5691$_{\pm 0.018}$ &0.5069  \\
&ComGA (WSDM'22) &0.5895$_{\pm 0.008}$ & 0.6055$_{\pm 0.000}$ & 0.5453$_{\pm 0.003}$ & 0.4391$_{\pm 0.000}$  & 0.7154$_{\pm 0.014}$  &0.5352$_{\pm 0.006}$ &  0.5716\\
&TAM (NeurIPS'23)   &\underline{0.7064}$_{\pm 0.010}$  &\underline{0.9144}$_{\pm 0.008}$  &0.6023$_{\pm 0.004}$   &\underline{0.5643}$_{\pm 0.007}$ &\underline{0.8476}$_{\pm 0.028}$  &\underline{0.5818}$_{\pm 0.033}$ &\underline{0.7028} \\
\cline{2-9}
&UNPrompt (Ours) &\textbf{0.7335}$_{\pm 0.020}$&\textbf{0.9379}$_{\pm 0.006}$&\textbf{0.6067}$_{\pm 0.006}$&\textbf{0.6223}$_{\pm 0.007}$ &\textbf{0.8516}$_{\pm 0.004}$   &\textbf{0.6084}$_{\pm 0.001}$ &\textbf{0.7267}\\
\hline
\hline
\multirow{9}*{AUPRC}
&iForest (TKDD'12) &0.1371$_{\pm 0.002}$ &0.0316$_{\pm 0.003}$ &0.0269$_{\pm 0.001}$ &0.0409$_{\pm 0.000}$ &  0.0399$_{\pm 0.001}$   & 0.1092$_{\pm 0.001}$ &0.0643\\
&ANOMALOUS (IJCAI'18) &0.0558$_{\pm 0.001}$ &0.1898$_{\pm 0.004}$ &0.0375$_{\pm 0.004}$ &0.0519$_{\pm 0.002}$ &0.0321$_{\pm 0.001}$  & 0.0361$_{\pm 0.005}$&0.0672\\
&DOMINANT (SIAM'19) &0.1424$_{\pm 0.002}$  &0.0314$_{\pm 0.041}$  &0.0356$_{\pm 0.002}$  &0.0395$_{\pm 0.020}$ &0.1015$_{\pm 0.018}$   &0.1638$_{\pm 0.007}$&0.0857\\
&CoLA (TNNLS'21) &0.0677$_{\pm 0.001}$ &0.2106$_{\pm 0.017}$ &\underline{0.0449}$_{\pm 0.002}$ &0.0448$_{\pm 0.002}$ &   0.0516$_{\pm 0.001}$    &0.1361$_{\pm 0.015}$&0.0926\\
&SL-GAD (TKDE'21) & 0.0634$_{\pm 0.005}$  &0.1316$_{\pm 0.020}$  &0.0406$_{\pm 0.004}$  &0.0350$_{\pm 0.000}$   & 0.0444$_{\pm 0.001}$    &0.1711$_{\pm 0.011}$&0.0810\\
&HCM-A (ECML-PKDD'22) & 0.0527$_{\pm 0.015}$   & 0.0713$_{\pm 0.004}$  &0.0287$_{\pm 0.005}$  &0.0287$_{\pm 0.012}$ & 0.0565$_{\pm 0.003}$  &  0.1154$_{\pm 0.004}$&0.0589\\
&ComGA (WSDM'22) &0.1153$_{\pm 0.005}$  &0.0354$_{\pm 0.001}$  &0.0374$_{\pm 0.001}$  & 0.0423$_{\pm 0.000}$  &0.1854$_{\pm 0.003}$  &0.1658$_{\pm 0.003}$&0.0969 \\
&TAM (NeurIPS'23) &\underline{0.2634}$_{\pm 0.008}$  &\underline{0.2233}$_{\pm 0.016}$  &0.0446$_{\pm 0.001}$  &\underline{0.0778}$_{\pm 0.009}$   & \underline{0.4346}$_{\pm 0.021}$   &\underline{0.1886}$_{\pm 0.017}$&\underline{0.2054} \\
\cline{2-9}
&UNPrompt (Ours) &\textbf{0.2688}$_{\pm 0.060}$&\textbf{0.2622}$_{\pm 0.028}$&\textbf{0.0450}$_{\pm 0.001}$&\textbf{0.0895}$_{\pm 0.004}$ &\textbf{0.6094}$_{\pm 0.014}$  &\textbf{0.2068}$_{\pm 0.004}$&\textbf{0.2470}\\
\hline
\bottomrule
\end{tabular}}
\end{center}
\caption{AUROC and AUPRC results of unsupervised GAD methods on six real-world GAD datasets. The best performance per column within each metric is boldfaced, with the second-best underlined. ``Avg'' denotes the average performance of each method.}
\label{unsupervised2}
\end{table*}

\begin{figure*}[!ht]
\centering
\subfigure[Amazon]{\includegraphics[width=0.22\textwidth]{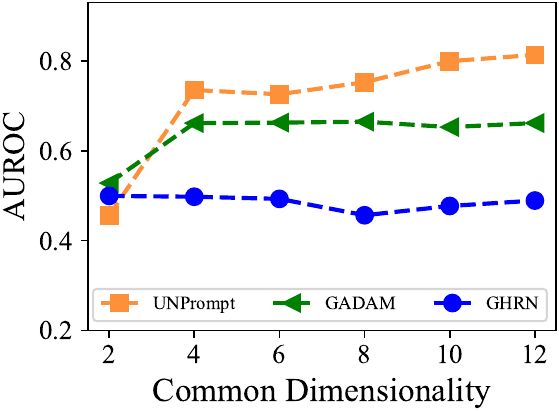}} 
\subfigure[Reddit]{\includegraphics[width=0.22\textwidth]{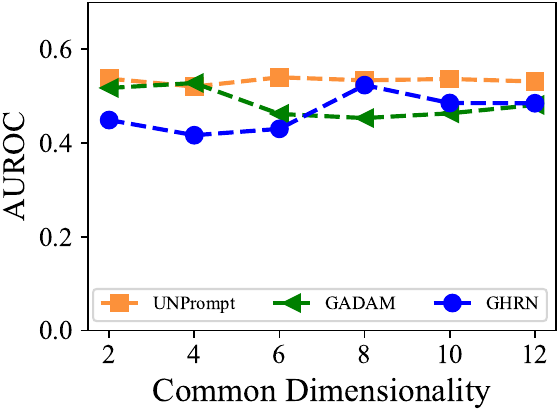}} 
\subfigure[Weibo]{\includegraphics[width=0.22\textwidth]{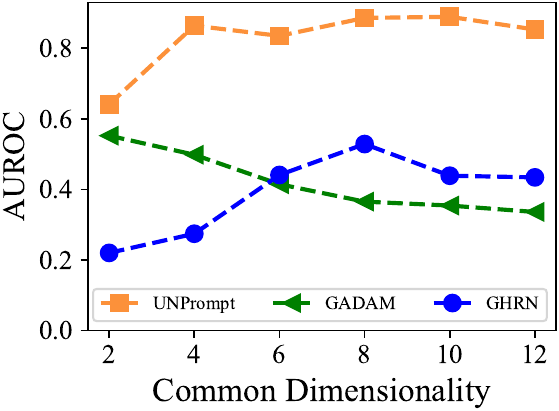}} 
\subfigure[YelpChi]{\includegraphics[width=0.22\textwidth]{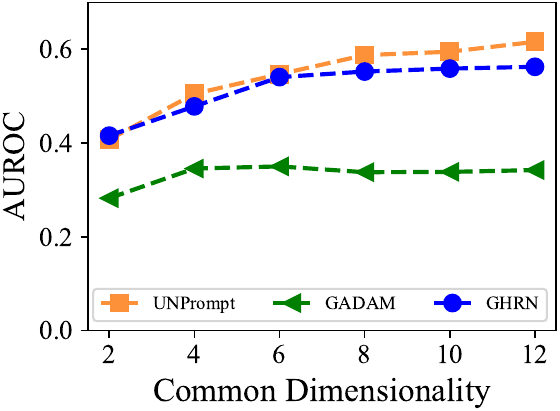}} 
\caption{AUROC results of UNPrompt and other baselines with different common dimensionalities.}
\label{per_vs_dim}
\end{figure*}

\subsection{Experimental Setup.}

Six datasets from different distributions and domains are used, \ie, Amazon, Facebook, Reddit, YelpChi, Amazon-all, and YelpChi-all. Following \cite{qiao2024truncated}, eight SotA unsupervised baselines are used for comparison, \ie, iForest \cite{liu2012isolation}, ANOMALOUS \cite{peng2018anomalous}, CoLA \cite{liu2021anomaly}, SL-GAD \cite{zheng2021generative}, HCM-A \cite{huang2022hop}, DOMINANT \cite{ding2019deep}, ComGA \cite{luo2022comga} and TAM \cite{qiao2024truncated}. For each method, we report the average performance with standard deviations after 5 independent runs with different random seeds. The implementation details of UNPrompt remain the same as in the generalist GAD setting. More experimental details on unsupervised GAD are in Sec.~\ref{exp_details}.

\subsection{Main Results.}
The AUROC and AUPRC results of all methods are presented in Table~\ref{unsupervised2}. Despite being a generalist GAD method, UNPrompt works very well as a specialized GAD model too. UNPrompt substantially outperforms all the competing methods on all datasets in terms of both AUROC and AUPRC. Particularly, the average performance of UNPrompt surpasses the best-competing method TAM by over 2\% in both metrics. Moreover, UNPrompt outperforms the best-competing method by  2\%-6\% in AUROC on most of the datasets. The superior performance shows that the latent node attribute predictability can be a generalized GAD measure that holds for different graphs, and this property can be effectively learned by the proposed neighborhood prompting method.

\section{{More Experimental Results}}
\subsection{{Generalist Performance with different common dimensionalities}}

For the results reported in the main paper, the common dimensionality is set to eight. In this subsection, we further evaluate the generalist anomaly detection with different common dimensionalities. Specifically, the dimensionality varies in $[2, 4, 6, 8, 10, 12]$ and the results are reported in Figure~\ref{per_vs_dim}. Without loss of generality, we conduct the analysis on Amazon, Reddit, Weibo and YelpChi and employ two baselines (GADAM and GHRN) for comparisons which are unsupervised and supervised methods, respectively.

From the figure, we can see that small dimensionality leads to poor generalist anomaly detection performance of UNPrompt. This is attributed to the fact that much attribute information would be discarded with a small dimensionality. By increasing the common dimensionality, more attribute information is retained, generally resulting in much better detection performance. However, this trend does not hold for the baselines as the baselines achieve inconsistent performance gain or loss on different datasets with the increase of the common dimensionality. This is largely attributed to the fact that these one-model-for-one-dataset methods fail to capture the generalized anomaly measure.

\subsection{{Incorporating coordinate-wise normalization into baselines}}
We further conduct experiments by incorporating the proposed coordinate-wise normalization into the baselines to evaluate whether the normalization could facilitate the baselines. Without loss of generality, three unsupervised methods (AnomalyDAE, CoLA and TAM) and three supervised methods (GCN, BWGNN and GHRN) are used and the results are reported in Table~\ref{baseline_wNor}.

\begin{table*}[t]
\begin{center}
\resizebox{0.90\textwidth}{!}{
\begin{tabular}{c|l|ccccccc|c}
\hline
\hline
\multirow{2}*{\textbf{Metric}}&\multirow{2}*{\textbf{Method}} & \multicolumn{7}{c|}{\textbf{Dataset}}\\
&&Amazon &Reddit &Weibo & YelpChi &Aamzon-all &YelpChi-all& Disney&Avg.\\
\hline
\multirow{15}{*}{AUROC}
&\multicolumn{9}{c}{\cellcolor[HTML]{EFEFEF}Unsupervised Methods}\\
&AnomalyDAE &0.5818$_{\pm 0.039}$&0.5016$_{\pm 0.032}$&\underline{0.7785}$_{\pm 0.058}$&0.4837$_{\pm 0.094}$&0.7228$_{\pm 0.023}$&0.5002$_{\pm 0.018}$&0.4853$_{\pm 0.003}$&\underline{0.5791}\\
&\quad + CN &0.4359$_{\pm 0.053}$&0.4858$_{\pm 0.063}$&0.4526$_{\pm 0.074}$&\textbf{0.5992$_{\pm 0.028}$}&0.2833$_{\pm 0.039}$&0.5080$_{\pm 0.013}$&0.5042$_{\pm 0.065}$&0.4670\\
&CoLA    &0.4580$_{\pm 0.054}$&0.4623$_{\pm 0.005}$&0.3924$_{\pm 0.041}$&0.4907$_{\pm 0.017}$&0.4091$_{\pm 0.052}$&0.4879$_{\pm 0.010}$&0.4696$_{\pm 0.065}$&0.4529\\
&\quad + CN &0.4729$_{\pm 0.019}$&0.5299$_{\pm 0.008}$&0.3401$_{\pm 0.026}$&0.3640$_{\pm 0.006}$&0.5424$_{\pm 0.019}$&0.4882$_{\pm 0.008}$&0.5593$_{\pm 0.079}$&0.4710\\
&TAM     &0.4720$_{\pm 0.005}$&\textbf{0.5725}$_{\pm 0.004}$&0.4867$_{\pm 0.028}$&0.5035$_{\pm 0.014}$&0.7543$_{\pm 0.002}$&0.4216$_{\pm 0.002}$&0.4773$_{\pm 0.003}$&0.5268\\
&\quad + CN &0.4509$_{\pm 0.015}$&0.5526$_{\pm 0.006}$&0.4723$_{\pm 0.007}$&0.5189$_{\pm 0.006}$&\underline{0.7580$_{\pm 0.004}$}&0.4057$_{\pm 0.002}$&0.2431$_{\pm 0.029}$&0.4859\\
&\multicolumn{9}{c}{\cellcolor[HTML]{EFEFEF}Supervised Methods}\\
&GCN    &\underline{0.5988}$_{\pm 0.016}$&\underline{0.5645}$_{\pm 0.000}$&0.2232$_{\pm 0.074}$&0.5366$_{\pm 0.019}$&0.7195$_{\pm 0.002}$&\underline{0.5486}$_{\pm 0.001}$&0.5000$_{\pm 0.000}$&0.5273\\
&\quad + CN &0.5694$_{\pm 0.014}$&0.5349$_{\pm 0.008}$&0.0632$_{\pm 0.005}$&0.3954$_{\pm 0.002}$&0.6798$_{\pm 0.009}$&0.5550$_{\pm 0.005}$&0.5507$_{\pm 0.015}$&0.4783\\
&BWGNN  &0.4769$_{\pm 0.020}$&0.5208$_{\pm 0.016}$&0.4815$_{\pm 0.108}$&0.5538$_{\pm 0.027}$&0.3648$_{\pm 0.050}$&0.5282$_{\pm 0.015}$&0.6073$_{\pm 0.026}$&0.5048\\
&\quad + CN &0.4745$_{\pm 0.048}$&0.4942$_{\pm 0.011}$&0.2538$_{\pm 0.038}$&0.4727$_{\pm 0.016}$&0.6307$_{\pm 0.077}$&0.5221$_{\pm 0.025}$&0.6042$_{\pm 0.039}$&0.4932\\
&GHRN   &0.4560$_{\pm 0.033}$&0.5253$_{\pm 0.006}$&0.5318$_{\pm 0.038}$&0.5524$_{\pm 0.020}$&0.3382$_{\pm 0.085}$&0.5125$_{\pm 0.016}$&0.5336$_{\pm 0.030}$&0.4928\\
&\quad + CN &0.4308$_{\pm 0.024}$&0.5061$_{\pm 0.026}$&0.2621$_{\pm 0.043}$&0.4781$_{\pm 0.018}$&0.5712$_{\pm 0.046}$&0.5200$_{\pm 0.009}$&\underline{0.6220}$_{\pm 0.032}$&0.4843\\
\cline{2-10}
&UNPrompt (Ours)    &\textbf{0.7525}$_{\pm 0.016}$&0.5337$_{\pm 0.002}$&\textbf{0.8860}$_{\pm 0.007}$&\underline{0.5875}$_{\pm 0.016}$&\textbf{0.7962}$_{\pm 0.022}$&\textbf{0.5558}$_{\pm 0.012}$&\textbf{0.6412}$_{\pm 0.030}$&\textbf{0.6790}\\
\hline
\hline
\multirow{15}{*}{AUPRC}
&\multicolumn{9}{c}{\cellcolor[HTML]{EFEFEF}Unsupervised Methods}\\
&AnomalyDAE &0.0833$_{\pm 0.015}$&0.0327$_{\pm 0.004}$&\underline{0.6064}$_{\pm 0.031}$&0.0624$_{\pm 0.017}$&0.1921$_{\pm 0.026}$&0.1484$_{\pm 0.009}$&0.0566$_{\pm 0.000}$&\underline{0.1688}\\
&\quad + CN &0.0596$_{\pm 0.009}$&0.0333$_{\pm 0.007}$&0.1910$_{\pm 0.049}$&\textbf{0.0874$_{\pm 0.011}$}&0.0495$_{\pm 0.006}$&0.1527$_{\pm 0.007}$&\underline{0.1232}$_{\pm 0.023}$&0.0995\\
&CoLA    &0.0669$_{\pm 0.002}$&0.0391$_{\pm 0.004}$&0.1189$_{\pm 0.014}$&0.0511$_{\pm 0.000}$&0.0861$_{\pm 0.019}$&0.1466$_{\pm 0.003}$&0.0701$_{\pm 0.023}$&0.0827\\
&\quad + CN &0.0669$_{\pm 0.002}$&0.0360$_{\pm 0.002}$&0.1618$_{\pm 0.027}$&0.0370$_{\pm 0.000}$&0.0934$_{\pm 0.017}$&0.1446$_{\pm 0.005}$&0.0870$_{\pm 0.025}$&0.0895\\
&TAM     &0.0666$_{\pm 0.001}$&\underline{0.0413}$_{\pm 0.001}$&0.1240$_{\pm 0.014}$&0.0524$_{\pm 0.002}$&{0.1736}$_{\pm 0.004}$&0.1240$_{\pm 0.001}$&0.0628$_{\pm 0.001}$&0.0921\\
&\quad + CN &0.0606$_{\pm 0.003}$&0.0394$_{\pm 0.001}$&0.1044$_{\pm 0.005}$&0.0542$_{\pm 0.001}$&\textbf{0.2482$_{\pm 0.013}$}&0.1213$_{\pm 0.001}$&0.0366$_{\pm 0.003}$&0.0950\\
&\multicolumn{9}{c}{\cellcolor[HTML]{EFEFEF}Supervised Methods}\\
&GCN     &\underline{0.0891}$_{\pm 0.007}$&\textbf{0.0439}$_{\pm 0.000}$&0.1109$_{\pm 0.020}$&0.0648$_{\pm 0.009}$&0.1536$_{\pm 0.002}$&\underline{0.1735}$_{\pm 0.000}$&0.0484$_{\pm 0.000}$&0.0977\\
&\quad + CN &0.0770$_{\pm 0.003}$&0.0355$_{\pm 0.001}$&0.0548$_{\pm 0.000}$&0.0401$_{\pm 0.000}$&0.1383$_{\pm 0.006}$&0.1789$_{\pm 0.002}$&0.0968$_{\pm 0.020}$&0.0888\\
&BWGNN   &0.0652$_{\pm 0.002}$&0.0389$_{\pm 0.003}$&0.2241$_{\pm 0.046}$&0.0708$_{\pm 0.018}$&0.0586$_{\pm 0.003}$&0.1605$_{\pm 0.005}$&0.0624$_{\pm 0.003}$&0.0972\\
&\quad + CN &0.0684$_{\pm 0.014}$&0.0320$_{\pm 0.001}$&0.2576$_{\pm 0.031}$&0.0516$_{\pm 0.004}$&0.1557$_{\pm 0.115}$&0.1585$_{\pm 0.010}$&0.0975$_{\pm 0.016}$&0.1173\\
&GHRN    &0.0633$_{\pm 0.003}$&0.0407$_{\pm 0.002}$&0.1965$_{\pm 0.059}$&0.0661$_{\pm 0.010}$&0.0569$_{\pm 0.006}$&0.1505$_{\pm 0.005}$&0.0519$_{\pm 0.003}$&0.0894\\
&\quad + CN &0.0586$_{\pm 0.004}$&0.0330$_{\pm 0.002}$&0.2663$_{\pm 0.038}$&0.0525$_{\pm 0.004}$&0.0898$_{\pm 0.015}$&0.1570$_{\pm 0.007}$&0.1051$_{\pm 0.021}$&0.1089\\
\cline{2-10}
&UNPrompt (Ours)     &\textbf{0.1602}$_{\pm 0.013}$&0.0351$_{\pm 0.000}$&\textbf{0.6406}$_{\pm 0.026}$&\underline{0.0712}$_{\pm 0.008}$&\underline{0.2430}$_{\pm 0.028}$&\textbf{0.1810}$_{\pm 0.012}$&\textbf{0.1236}$_{\pm 0.031}$&\textbf{0.2078}\\
\hline 
\hline
\end{tabular}
}
\end{center}
\caption{{AUROC and AUPRC results of several baselines with coordinate-wise normalization (CN).}}
\label{baseline_wNor}
\end{table*}

From the table, we can see that the proposed coordinate-wise normalization does not improve the baselines consistently but downgrades most of the baselines. This can be attributed to two reasons. First, while the proposed coordinate-wise normalization unifies the semantics of different graphs into a common space, the discrimination between normal and abnormal patterns would also be compressed. This requires the generalist anomaly detector to capture the fine-grained differences between normal and abnormal patterns. Second, these baselines are not designed to capture generalized abnormality and normality across graphs, failing to capture and discriminate the generalized nuance. By contrast, we reveal that the predictability of latent node attributes can serve as a generalized anomaly measure and learn highly generalized normal and abnormal patterns via latent node attribute prediction. In this way, the graph-agnostic anomaly measure can be well generalized across graphs.

\subsection{{Generalist performance with different training graphs}}
In the main paper, we report the generalist performance of UNPrompt by using Facebook as the training graph. To further demonstrate the generalizability of UNPrompt, we conduct additional experiments by using Amazon as the training graph and testing the learned generalist model on the rest graphs. Note that Facebook and Amazon are from different domains, which are the social network and the co-review network respectively.

The AUROC and AUPRC results of all methods are reported in Table~\ref{gen_auroc_amzon}. Similar to the observations when taking Facebook as the training graph, UNPrompt achieves the best average performance in terms of both AUROC and AUPRC when training on Amazon, demonstrating the generalizability and effectiveness of UNPrompt with different training graphs. Note that the training graph Amazon and the target graph Amazon-all come from the same distribution but have different numbers of nodes and graph structures. Intuitively, all the methods should achieve promising performance on Amazon-all. However, only a few methods achieve this goal, including BWGNN, GHRN, XGBGraph, and our method. The failures of other baselines can be attributed to the more complex graph structure of Amazon-all hinders the generalizability of these methods. Moreover, compared to BWGNN, GHRN and XGBGraph, our method performs more stably across different datasets. This demonstrates the importance of capturing intrinsic normal and abnormal patterns for graph anomaly detection.

\begin{table*}[t]
\begin{center}
\resizebox{0.9\textwidth}{!}{
\begin{tabular}{c|c|ccccccc|c}
\hline
\hline
\multirow{2}*{\textbf{Metric}}&\multirow{2}*{\textbf{Method}} & \multicolumn{7}{c|}{\textbf{Dataset}}\\
&&Facebook &Reddit &Weibo & YelpChi &Aamzon-all &YelpChi-all&Disney&Avg.\\
\hline
\multirow{13}{*}{AUROC}
&\multicolumn{9}{c}{\cellcolor[HTML]{EFEFEF}Unsupervised Methods}\\
&AnomalyDAE &0.6123$_{\pm 0.141}$&\textbf{0.5799}$_{\pm 0.035}$&\underline{0.7884}$_{\pm 0.031}$&0.4788$_{\pm 0.046}$&0.6233$_{\pm 0.070}$&0.4912$_{\pm 0.009}$&0.4938$_{\pm 0.005}$&0.5811\\
&CoLA    &0.5427$_{\pm 0.109}$&0.4962$_{\pm 0.025}$&0.3987$_{\pm 0.017}$&0.3358$_{\pm 0.012}$&0.4751$_{\pm 0.014}$&0.4937$_{\pm 0.003}$&0.5455$_{\pm 0.031}$&0.4697\\
&HCM-A   &0.5044$_{\pm 0.047}$&0.4993$_{\pm 0.057}$&0.4937$_{\pm 0.056}$&0.5000$_{\pm 0.000}$&0.4785$_{\pm 0.016}$&0.4958$_{\pm 0.003}$&0.2051$_{\pm 0.034}$&0.4538\\
&TAM     &0.5496$_{\pm 0.038}$&\underline{0.5764}$_{\pm 0.003}$&0.4876$_{\pm 0.029}$&0.5091$_{\pm 0.014}$&0.7525$_{\pm 0.002}$&0.4268$_{\pm 0.002}$&0.4850$_{\pm 0.004}$&0.5410\\
&GADAM &0.6024$_{\pm 0.033}$&0.4720$_{\pm 0.062}$&0.4324$_{\pm 0.047}$&0.4299$_{\pm 0.023}$&0.5199$_{\pm 0.072}$&0.5289$_{\pm 0.017}$&0.3966$_{\pm 0.021}$&0.4832\\
&\multicolumn{9}{c}{\cellcolor[HTML]{EFEFEF}Supervised Methods}\\
&GCN    &\underline{0.6892}$_{\pm 0.004}$&0.5658$_{\pm 0.000}$&0.2355$_{\pm 0.019}$& 0.5277$_{\pm 0.002}$&0.7503$_{\pm 0.002}$&0.5565$_{\pm 0.000}$&0.5000$_{\pm 0.000}$&0.5464\\
&GAT    &0.3886$_{\pm 0.118}$&0.4997$_{\pm 0.012}$&0.3897$_{\pm 0.134}$&0.5051$_{\pm 0.019}$&0.5007$_{\pm 0.006}$&0.4977$_{\pm 0.006}$&0.5840$_{\pm 0.111}$&0.4808\\
&BWGNN  &0.5441$_{\pm 0.020}$&0.4026$_{\pm 0.028}$&0.4214$_{\pm 0.039}$&0.4908$_{\pm 0.013}$&\underline{0.9684}$_{\pm 0.005}$&0.5841$_{\pm 0.062}$&0.4196$_{\pm 0.047}$&0.5473\\
&GHRN   &0.5242$_{\pm 0.013}$&0.4096$_{\pm 0.021}$&0.4783$_{\pm 0.021}$&0.5036$_{\pm 0.016}$&0.9601$_{\pm 0.018}$&\textbf{0.6045}$_{\pm 0.022}$&0.4000$_{\pm 0.098}$&0.5543\\
&XGBGraph &0.4869$_{\pm 0.069}$&0.4869$_{\pm 0.069}$&0.7843$_{\pm 0.090}$&0.4773$_{\pm 0.022}$&\textbf{0.9815$_{\pm 0.000}$}&\underline{0.5869}$_{\pm 0.014}$&0.4376$_{\pm 0.044}$&\underline{0.6059}\\
&GraphPrompt &0.3093$_{\pm 0.000}$&0.4511$_{\pm 0.000}$&0.2032$_{\pm 0.000}$&\textbf{0.7093}$_{\pm 0.000}$&0.6331$_{\pm 0.000}$&0.4994$_{\pm 0.000}$&\textbf{0.7128}$_{\pm 0.000}$&0.5026\\
\cline{2-10}
&Our    &\textbf{0.7917}$_{\pm 0.021}$&0.5356$_{\pm 0.005}$&\textbf{0.8192}$_{\pm 0.015}$&\underline{0.5362}$_{\pm 0.007}$&0.9289$_{\pm 0.007}$&0.5448$_{\pm 0.009}$&\underline{0.6959}$_{\pm 0.042}$&\textbf{0.6932}\\
\hline
\hline
\multirow{13}{*}{AUPRC}
&\multicolumn{9}{c}{\cellcolor[HTML]{EFEFEF}Unsupervised Methods}\\
&AnomalyDAE     &\underline{0.0675}$_{\pm 0.028}$&0.0413$_{\pm 0.005}$&\textbf{0.6172}$_{\pm 0.015}$&0.0647$_{\pm 0.016}$&0.1025$_{\pm 0.026}$&0.1479$_{\pm 0.006}$&0.0583$_{\pm 0.001}$&0.1571\\
&CoLA    &0.0468$_{\pm 0.026}$&0.0327$_{\pm 0.002}$&0.0956$_{\pm 0.005}$&0.0361$_{\pm 0.001}$&0.0678$_{\pm 0.005}$&0.1474$_{\pm 0.001}$&0.0717$_{\pm 0.015}$&0.0712\\
&HCM-A   &0.0249$_{\pm 0.003}$&0.0374$_{\pm 0.008}$&0.0979$_{\pm 0.011}$&0.0511$_{\pm 0.000}$&0.0727$_{\pm 0.006}$&0.1453$_{\pm 0.000}$&0.0452$_{\pm 0.020}$&0.0678\\
&TAM     &0.0243$_{\pm 0.002}$&\underline{0.0417}$_{\pm 0.001}$&0.1266$_{\pm 0.015}$&0.0532$_{\pm 0.002}$&0.1771$_{\pm 0.002}$&0.1271$_{\pm 0.001}$&0.0682$_{\pm 0.002}$&0.0883\\
&GADAM &0.0461$_{\pm 0.014}$&0.0299$_{\pm 0.004}$&0.0917$_{\pm 0.007}$&0.0428$_{\pm 0.002}$&0.0773$_{\pm 0.024}$&0.1602$_{\pm 0.010}$&0.0732$_{\pm 0.004}$&0.0745\\
&\multicolumn{9}{c}{\cellcolor[HTML]{EFEFEF}Supervised Methods}\\
&GCN     &0.0437$_{\pm 0.001}$&\textbf{0.0449}$_{\pm 0.000}$&0.2527$_{\pm 0.026}$&\underline{0.0763}$_{\pm 0.001}$&0.1738$_{\pm 0.002}$&0.1759$_{\pm 0.000}$&0.0484$_{\pm 0.000}$&0.1165\\
&GAT     &0.0445$_{\pm 0.039}$&0.0327$_{\pm 0.001}$&0.0892$_{\pm 0.016}$&0.0595$_{\pm 0.003}$&0.0697$_{\pm 0.001}$&0.1478$_{\pm 0.003}$&0.0760$_{\pm 0.035}$&0.0742\\
&BWGNN   &0.0289$_{\pm 0.003}$&0.0263$_{\pm 0.002}$&0.2735$_{\pm 0.026}$&0.0543$_{\pm 0.004}$&\underline{0.8406}$_{\pm 0.012}$&0.1975$_{\pm 0.031}$&0.0494$_{\pm 0.004}$&0.2101\\
&GHRN    &0.0254$_{\pm 0.001}$&0.0265$_{\pm 0.002}$&0.3103$_{\pm 0.013}$&0.0541$_{\pm 0.005}$&0.8142$_{\pm 0.045}$&\textbf{0.2015}$_{\pm 0.015}$&0.0561$_{\pm 0.028}$&0.2126\\
&XGBGraph &0.0268$_{\pm 0.006}$&0.0315$_{\pm 0.000}$&0.4116$_{\pm 0.040}$&0.0500$_{\pm 0.003}$&\textbf{0.8673}$_{\pm 0.000}$&\underline{0.1994}$_{\pm 0.012}$&0.0541$_{\pm 0.005}$&\underline{0.2344}\\
&GraphPrompt &0.0169$_{\pm 0.000}$&0.0298$_{\pm 0.000}$&0.0812$_{\pm 0.000}$&\textbf{0.0975}$_{\pm 0.000}$&0.1368$_{\pm 0.000}$&0.1481$_{\pm 0.000}$&\textbf{0.1157}$_{\pm 0.000}$&0.0894\\
\cline{2-10}
&Our     &\textbf{0.2291}$_{\pm 0.023}$&0.0340$_{\pm 0.001}$&\underline{0.4746}$_{\pm 0.033}$&0.0610$_{\pm 0.003}$&0.7329$_{\pm 0.042}$&0.1767$_{\pm 0.004}$&\underline{0.0933}$_{\pm 0.013}$&\textbf{0.2574}\\
\hline 
\hline
\end{tabular}
}
\end{center}
\caption{{AUROC and AUPRC results on seven real-world GAD datasets with the generalist model trained on Amazon. For each dataset and metric, the best performance per column is boldfaced, with the second-best underlined. ``Avg'' denotes the average performance of each method.}}
\label{gen_auroc_amzon}
\end{table*}

\subsection{Sensitivity of Unsuperivised GAD w.r.t the threshold}

To evaluate the sensitivity of the unsupervised GAD performance of our method w.r.t the threshold, we vary the threshold in the range of $[5\%, 50\%]$ at a step size of 5\%. Without loss of generality, three datasets are employed, including Facebook, Reddit, and Amazon. The results of these datasets are reported in the following table.

\begin{table}[ht]
\centering
\resizebox{0.5\textwidth}{!}{
\begin{tabular}{c|cccccccccc}
\hline
Datasets	&5	&10	&15	&20	&25	&30	&35	&40	&45	&50\\
\hline
Facebook&	0.8859	&0.9125	&0.9218	&0.9231	&0.9210	&0.9272	&0.9340	&0.9379	&0.9369	&0.9255\\
Reddit	&0.5667	&0.5754	&0.5802	&0.5824	&0.5942	&0.5871&	0.5864	&0.6067	&0.5906 &	0.5828\\
Amazon	&0.6264	&0.6960	&0.7152	&0.7393	&0.7394	&0.7463&	0.7462	&0.7335	&0.7392	&0.7389  \\
\hline
\end{tabular}}
\caption{AUROC results of UNPrompt with different thresholds (\%) in unsupervised GAD.}
\label{thres}
\end{table}

From the table, we can see that the unsupervised performance of UNPrompt can perform well and stably with a sufficiently large threshold (e.g., no less than 30\%), but it may drop significantly with a small threshold, e.g., thresholds like 5\%-15\% that are close to the ground-truth anomaly rate. This is because more abnormal nodes would have a substantially higher chance of being mistakenly treated as normal nodes with such a small threshold, which would in turn mislead the optimization and subsequently degrade the GAD performance. By contrast, increasing the threshold would lift the acceptance bar of normal nodes, allowing the optimization to focus on high-confident normal nodes and effectively mitigate the adverse effects caused by the wrongly labeled normal nodes.

\section{Time Complexity Analysis of UNPrompt}\label{timecomplexity}

\paragraph{Theoretical Analysis.} In this section, we analyze the time complexity of training UNPrompt. As discussed in the main paper, UNPrompt first pre-trains the aggregation network with graph contrastive learning. Then, the model remains frozen when optimizing neighborhood graph prompts and the transformation layer to capture the generalized normal and abnormal graph patterns. In the experimental section, we employ a one-layer aggregation network, denoting the number of hidden units as $d_h$. The time complexity of the graph contrastive learning is $\mathcal{O}(4E_1(|A|d_h+Nd_hd^{'}+6Nd_h^2))$, where $|A|$ returns of the number of edges of the $\mathcal{G}_{\text{train}}$, $N$ is the number of nodes, $d^{'}$ represents the predefined dimensionality of node attributes, and $E_1$ is the number of training epoch. After that, we freeze the learned model and learn the learnable neighborhood prompt tokens and the transformation layer to capture the shared anomaly patterns. In our experiments, we set the size of each graph prompt to $K$ and implement the classification head as a single-layer MLP with the same hidden units $d_h$. Given the number of the training epoch $E_2$, the time complexity of optimizing the graph prompt and the transformation layer is $\mathcal{O}((4KNd^{'}+2Nd_h^2)E_2)$, which includes both the forward and backward propagation. Note that, despite the neighborhood aggregation model being frozen, the forward and backward propagations of the model are still needed to optimize the task-specific graph prompts and the transformation layer. Therefore, the overall time complexity of UNPrompt is $\mathcal{O}(4E_1(|A|d_h+Nd_hd^{'}+6Nd_h^2) + 2E_2(|A|d_h+Nd_hd^{'} + 2KNd^{'}+Nd_h^2))$, which is linear to the number of nodes, the number of edges, and the number of node attributes of the training graph. Note that, after the training, the learned generalist model is directly utilized to perform anomaly detection on various target graphs without any further training.

\begin{table}[ht]
\centering
\begin{center}
\resizebox{0.5\textwidth}{!}{
\begin{tabular}{c|ccccc}
\hline
Methods&  AnomalyDAE & TAM & GAT & BWGNN & UNPrompt\\
\hline
Training Time  &86.04&479.70&2.43&4.86&2.08\\
Inference Time &264.29&521.92&300.90&330.99&58.95\\
\hline
\end{tabular}}
\end{center}
\caption{Training time and inference time (seconds) for different methods.}
\label{time}
\end{table}

\paragraph{Empirical Analysis.}
In Table~\ref{time}, we report the training time and inference time of different methods, where two representative unsupervised methods (AnomalyDAE and TAM) and two supervised methods (GAT and BWGNN) are used for comparison to our method UNPrompt. The results show that the proposed method requires much less training and inference time compared to other baselines, demonstrating the efficiency of the proposed UNPrompt. Note that, TAM has the highest time consumption, which can be attributed to that it performs multiple graph truncation and learns multiple local affinity maximization networks.

\section{Experimental Setup}

\subsection{Details on Datasets}\label{dataset_detail}
We conduct the experiments using eight real-world with genuine anomalies in diverse online shopping services, social networks and co-purchase networks, including Facebook \cite{xu2022contrastive}, Reddit \cite{kumar2019predicting},  Weibo \cite{zhao2020error},  Amazon \cite{mcauley2013amateurs}, YelpChi \cite{rayana2015collective} as well as two large-scale graph datasets including Amazon-all \cite{mcauley2013amateurs} and YelpChi-all  \cite{rayana2015collective} and Disney \cite{sanchez2013statistical}. The statistical information, including the number of nodes, edges, the dimension of the feature, and the anomaly rate of the datasets can be found in Table~\ref{ref:datastats}. 

\begin{table}
\begin{center}
\resizebox{0.5\textwidth}{!}{
\begin{tabular}{lccccc}
\toprule
Data set & Type & Nodes & Edges & Attributes & Anomalies(Rate)  \\
\midrule
Facebook  & Social Networks& 1,081 &55,104 &576 &25(2.31\%)\\
Reddit  & Social Networks &10,984 &168,016 &64 &366(3.33\%)\\
Weibo & Social Networks &8,405 &407,963 &400 &868(10.30\%)\\
Amazon   & Co-review   &10,244 &175,608  &25 & 693(6.66\%)\\
YelpChi  & Co-review  &24,741 &49,315 &32 &1,217(4.91\%)\\
Amazon-all & Co-review    & 11,944 &  4,398,392&25 & 821(6.87\%)\\
YelpChi-all  & Co-review &45,941 &3,846,979 &32 &6,674(14.52\%)\\
Disney  & Co-purchase &124 & 335 & 28 & 6(4.84\%)\\
\bottomrule
\end{tabular}}
\end{center}
\caption{Key statistics of the real-world GAD datasets with real anomalies.}
\label{ref:datastats}
\end{table}

\subsection{More Implementation Details}\label{exp_details}
\paragraph{Generalist GAD.}
For the graph contrastive learning-based pre-training, the probabilities of edge removal and attribute masking are by default set to 0.2 and 0.3 respectively. Besides, the learning rate is set to 0.001 with the Adam optimizer, the training epoch is set to 200, and the temperature $\tau$ is 0.5. 

For the neighborhood prompt learning, the learning rate is also set to 0.001 with the Adam optimizer, and the training epoch is set to 900. Note that, since we focus on generalist GAD, we do not perform any hyperparameter search for specific target graphs. Instead, the results of all target graphs are obtained with the same hyperparameter settings.

\paragraph{Unsupervised GAD.}
Similar to the generalist GAD setting, the hidden units of the neighborhood aggregation network and the transformation layer are set to 128 for all graphs. The threshold $t$ is determined by the 40th percentile of the normal scores obtained by the pre-trained model $g$, and the scaling parameter $\alpha$ is set to 10 for all graphs. Besides, we utilize random search to find the optimal hyperparameters of the size of neighborhood prompts $K$ and the trade-off parameter $\lambda$.

For both generalist and unsupervised GAD, the code is implemented with Pytorch (version: 1.13.1), DGL (version: 1.0.1), OGB (version: 1.3.6), and Python 3.8.19. All experiments are conducted on a Linux server with an Intel CPU (Intel Xeon Gold 6346 3.1GHz) and an Nvidia A40 GPU.

\section{More Discussions on UNPrompt}

\subsection{Difference between Inductive learning and Generalist GAD}
Similar to the generalist setting, inductive graph learning \cite{hamilton2017inductive,ding2021inductive,li2023diga,huang2023unsupervised,fang2023anonymous} also focuses on inference on unseen graph data. However, these methods are not applicable to the generalist setting. Specifically, inductive graph learning trains the model on partial data of the whole graph dataset \cite{hamilton2017inductive,ding2019deep,li2023diga} or the previously observed data of dynamic graphs \cite{fang2023anonymous}. Then, the learned model is evaluated on the unseen data of the whole dataset or the future graph. These unseen testing data are from the same source of the training data with the same dimensionality and semantics. In contrast, the unseen data in our method are from different distributions/domains with significantly different dimensionality and semantics. This cross-dataset nature, specifically referred to as a zero-shot problem \cite{jeong2023winclip,zhou2024anomalyclip}, makes our setting significantly different from the current inductive graph learning setting. 

\subsection{Difference between ARC and UNPrompt}
While both ARC \cite{liu2024arc} and our UNPrompt focus on generalist graph anomaly detection, there exist significant differences between them. First, ARC addresses a few-shot setting, whereas UNPrompt addresses a zero-shot setting. Second, ARC performs anomaly scoring using a traditional anomaly measure, reconstruction error. It calculates the reconstruction errors of query node embeddings w.r.t the in-context embeddings of labeled nodes. Differently, UNPrompt proposes a novel generalized anomaly measure based on latent attribute predictability. Third, the training datasets for ARC consist of the largest graphs from all multiple graph types/domains. The test graphs from the same types/domains are used during inference. By contrast, UNPrompt requires only one single dataset to effectively learn the generalist model, and the testing datasets are from unseen datasets and domains. 

\subsection{Future works}
Despite the proposed method achieving promising performance for generalist GAD, there are several interesting future works that can be explored to further enhance the contribution of UNPrompt. First, like most existing methods, UNPrompt cannot process very large graphs with millions of nodes, as it requires loading the full graph during inference. A subgraph-based inference may be a solution, and we will investigate this problem in future work. Second, we assume that normal nodes can be well predicted by their neighbors, and thus we focus on their one-hop neighbors. The experiments verify the effectiveness of this assumption. However, it is still worth exploring the utilization of multi-hop neighbors. Third, it would be interesting and practical to extend the proposed method to dynamic graph anomaly detection \cite{jiang2024spade+,chen2024rush,jiang2022spade,liu2021anomalyd}. 

\end{document}